\documentclass[lettersize,journal]{IEEEtran}
\usepackage{amsmath,amsfonts}
\usepackage{algorithmic}
\usepackage{algorithm}
\usepackage{array}
\usepackage[caption=false,font=normalsize,labelfont=sf,textfont=sf]{subfig}
\usepackage{textcomp}
\newtheorem{theorem}{Theorem}[section]
\newtheorem{lemma}{Lemma}[section]
\newtheorem{definition}{Definition}[section]
\newtheorem{corollary}{Corollary}[section]
\newtheorem{proposition}{Proposition}[section]

\usepackage{stfloats}
\usepackage{url}
\usepackage{verbatim}
\usepackage{graphicx}
\usepackage{xcolor}
\usepackage{booktabs}
\usepackage{cite}
\usepackage{colortbl}
\makeatletter
\let\NAT@parse\undefined
\makeatother
\usepackage{hyperref}  
\hypersetup{
    colorlinks=true,
    linkcolor=blue,
    filecolor=blue,      
    urlcolor=blue,
    citecolor=cyan,
}

\begin{document}

\title{Learn A Flexible Exploration Model for Parameterized Action Markov Decision Processes}

\author{Zijian Wang, Bin Wang, Mingwen Shao,~\IEEEmembership{Member,~IEEE,} Hongbo Dou, Boxiang Tao        
\thanks{These authors are with China University of Petroleum(East China), Qingdao 266580, China.}
}

\markboth{Journal of \LaTeX\ IEEE trans,~Vol., No., November~2024}%
{Shell \MakeLowercase{\textit{et al.}}: A Sample Article Using IEEEtran.cls for IEEE Journals}


\maketitle

\begin{abstract}
Hybrid action models are widely considered an effective approach to reinforcement learning (RL) modeling. The current mainstream method is to train agents under Parameterized Action Markov Decision Processes (PAMDPs), which performs well in specific environments. Unfortunately, these models either exhibit drastic low learning efficiency in complex PAMDPs or lose crucial information in the conversion between raw space and latent space. To enhance the learning efficiency and asymptotic performance of the agent, we propose a model-based RL (MBRL) algorithm, FLEXplore. FLEXplore learns a parameterized-action-conditioned dynamics model and employs a modified Model Predictive Path Integral control. Unlike conventional MBRL algorithms, we carefully design the dynamics loss function and reward smoothing process to learn a loose yet flexible model. Additionally, we use the variational lower bound to maximize the mutual information between the state and the hybrid action, enhancing the exploration effectiveness of the agent. We theoretically demonstrate that FLEXplore can reduce the regret of the rollout trajectory through the Wasserstein Metric under given Lipschitz conditions. Our empirical results on several standard benchmarks show that FLEXplore has outstanding learning efficiency and asymptotic performance compared to other baselines.
\end{abstract}

\begin{IEEEkeywords}
Model-based reinforcement learning (MBRL), parameterized action Markov decision processes, Wasserstein Metric, Lipschitz analysis.
\end{IEEEkeywords}

\section{Introduction}
\IEEEPARstart{R}{ecent} years, reinforcement learning (RL) has achieved eye-catching success across various real-world domains, such as game playing \cite{mnih2015human}, robot control \cite{lillicrap2015continuous,schulman2015trust}, autonomous driving \cite{sallab2017deep}, because it mirrors the learning processes observed in human evolution. Continuous control represents a complex scenario, as it requires the agent to perform precise actions to accomplish tasks, placing greater demands on the design of the algorithm \cite{rezaei2022continuous,hansen2023td,yang2024movie,zhu2021making}.

Now, imagine you are playing a soccer game. When playing on the attacking side, you not only need to consider choices related to passing, dribbling, and shooting but also need to consider a range of continuous parameters such as power and angle. We abstract this issue, specifically focusing on the formulation of policies within the hybrid action space \cite{hu2021hierarchical,hausknecht2015deep,fu2019deep}. Evidently, this problem is inherently more complex than that of continuous control. Each discrete action may corresponds to distinct continuous action parameters, such as the speed parameters relevant to dribbling, which are not applicable during shooting. Compared to just discrete or continuous action space, Reinforcement Learning with Parameterized action space allows the agent to perform more structural exploration and solve more complex tasks with a semantically more meaningful action space \cite{hausknecht2015deep}. In the current RL community, an effective approach to addressing the challenges associated with hybrid action spaces is to establish the Parameterized Action Markov Decision Processes (PAMDPs) \cite{xiong2018parametrized,fan2019hybrid,li2021hyar}. For example, Hybrid PPO (HPPO) \cite{fan2019hybrid} uses multiple policy heads consisting of one for discrete actions and the others for corresponding continuous parameter of each discrete action separately.

These methods we mention above are model-free, which implicates that they will be trapped by the low sample efficiency. In relatively uncomplicated environments (for example, \texttt{gym\_platform} and \texttt{gym\_goal} \cite{Masson2016ParamActions}), these methods can still facilitate learning through adequate interaction with the environment. But in other complex environments, it is not realistic to complete the learning only by inefficient sampling \cite{li2021hyar}. By contrast, in continuous/discrete-only action spaces, deep model-based reinforcement learning (MBRL) has shown better performance than model-free approaches in many complex domains \cite{hafner2019dream, hafner2020mastering,hu2024privileged}. Therefore, we employ the MBRL manner: The agent learns a parameterized-action-conditioned dynamics model and employs a modified Model Predictive Path Integral control. To our best knowledge, DLPA \cite{zhang2024model} is the first approach integrating MBRL and PAMDPs, it learns the model by minimizing the L2 norm between the rollout states (and rewards) and the true state (and rewards). 

However, we observe several significant issues that are currently present:
\begin{itemize}
    \item \textbf{Issue 1.} In DLPA \cite{zhang2024model}, minimizing the L2 norm between learned model and environment dynamics may learn an overly precise model, undermining the exploration capabilities of the agent.\cite{young2022benefits, palenicek2023diminishing, babaeizadeh2020models, lee2023dreamsmooth}.

    \item \textbf{Issue 2.} Although most algorithms employ a stochastic policy network, they may converge to a nearly deterministic policy later in training, which makes them are prone to be trapped in local minima.

    \item \textbf{Issue 3.} Previous works have designed various neural networks to address the dependencies between discrete and continuous actions \cite{bester2019multi, zhang2022learning}. However, a clear mathematical criterion to guide subsequent actions is lacking, specifically regarding the extent to which discrete actions should influence continuous parameters in particular environments.

    \item \textbf{Issue 4.} Although DLPA \cite{zhang2024model} gives an important theorem for the regret of the rollout trajectory, it does not give a practical scheme to reduce the regret. This influences the asymptotic performance of the agent, which is the part we are most concerned about.
\end{itemize}

In light of these issues and challenges, we propose FLEXplore, an MBRL algorithm for PAMDPs. For \textbf{Issue 1}, we carefully design the loss function for dynamics learning within the framework of Lipschitz-PAMDPs \cite{zhang2024model} to learn a loose yet flexible dynamics. Inspired by \cite{asadi2018lipschitz}, we theoretically reveal the core advantange of Wasserstein Metric \cite{vaserstein1969markov} over other distribution metrics and the relationship between the loss function and the Wasserstein Metric between the learned dynamics and the environment dynamics. For \textbf{Issue 2}, we consider the effect of slight state shifts under the condition of fixed hybrid actions, that is, whether the agent can learn from other effective states in the vicinity of a highly rewarding state. Therefore, we employ the Fast Gradient Symbol Method (FGSM) \cite{goodfellow2014explaining} to design a reward smoothing mechanism \cite{lee2023dreamsmooth} for FLEXplore, which enables FLEXplore to learn additional states in the vicinity of high-reward states, facilitating escape from local minima while reducing the local Lipschitz constant of the reward function. For \textbf{Issue 3}, we introduce mutual information \cite{shannon1948mathematical} to assess the agent's exploration efficiency and improve its exploration by maximizing the variational lower-bound approximation \cite{barber2004algorithm} to increase the mutual information between the state and the hybrid action. By reformulating and relaxing the mutual information, we derive the dependence that should exist between the discrete action and the continuous parameter: The conditional entropy of the continuous parameter can be increased, while the conditional entropy of the induced state should be reduced. For \textbf{Issue 4}, we use the theorem presented in DLPA \cite{zhang2024model} to demonstrate that FLEXplore contributes to a reduction in the regret of the trajectory rollout. Our empirical results on 6 different PAMDPs benchmarks show that FLEXplore has outstanding learning efficiency and asymptotic performance compared to
other baselines. 

The contibution of our work are summarized as follows: (1) We carefully design a loss function to learn a losse yet flexible dynamics and theoretically reveal the relationship between this loss function and the Wasserstein Metric of learned dynamics and environment dynamics. (2) We propose FLEXplore, a MBRL algorithm for PAMDPs, which reduces the regret of the trajectory rollout through flexible model learning, reward smoothing, and mutual information maximizing. (3) Extensive experimental results demonstrate the superiority of our method in terms of both sample efficiency and asymptotic performance.

\section{Background}
\subsection{Lipschitz-PAMDPs}
Standard Markov Decision Processes (MDPs) defined by the tuple $(\mathcal{S},\mathcal{A},\mathcal{T},\mathcal{R},\gamma)$, where $\mathcal{S}$ and $\mathcal{A}$ are the state space and action space respectively, $\mathcal{T}$ denotes the state transition probability function, $\mathcal{R}$ is the reward function and $\gamma$ is the discount factor. In Parametrized Action Markov Decision Processes (PAMDPs), the action space $\mathcal{A}$ is divided into discrete action space $\mathcal{K}$ and its corresponding continuous parameter space $\mathcal{Z}$. For clarity, we denote the hybrid action space by $\mathcal{M} = \{ (k,z_k) | z_k \in \mathcal{Z}, k \in \{ 1,\dots,|\mathcal{K}| \} \}$, where $|\cdot|$ denotes the cardinal number. Our analyses leverage the "smoothness" of various functions, quantified as follows \cite{asadi2018lipschitz}.
\begin{definition}[Lipschitz Consistant]\label{def2.1}
    \textit{Given two metric spaces $(M_1,d_1)$ and $(M_2,d_2)$ consisting of a space and a distance metric, a function $f : M_1 \mapsto M_2$ is Lipschitz continuous if the Lipschitz constant, defined as}
    \begin{equation}\label{eq1}
        K_{d_1,d_2}(f) := \underset{s_1 \in M_1, s_2 \in M_2}{\text{sup}} \frac{d_2(f(s_1),f(s_2))}{d_1(s_1,s_2)},
    \end{equation}
    \textit{is \textit{finite}}.
\end{definition}
Based on \hyperref[def2.1]{Definition 2.1}, we can determine the dynamics $\mathcal{T}(\cdot|s,k,z_k)$ and reward function $\mathcal{R}(s,k,z_k)$ measures of their variables and obtain their corresponding Lipschitz constant $(L^\mathcal{S}_\mathcal{T},L^\mathcal{K}_\mathcal{T},L^\mathcal{Z}_\mathcal{T},L^\mathcal{S}_\mathcal{R},L^\mathcal{K}_\mathcal{R},L^\mathcal{Z}_\mathcal{R})$. Then we give the definition of Lipschitz-PAMDPs \cite{zhang2024model}.

\begin{definition}[Lipschitz-PAMDPs]
    A PAMDP is $(L^\mathcal{S}_\mathcal{T},L^\mathcal{K}_\mathcal{T},L^\mathcal{Z}_\mathcal{T})$-Lipschitz continuous if, for all $s \in \mathcal{S}$, $k \in \{ 1,\dots,|\mathcal{K}| \}$ and $z_k \in \mathcal{Z}$:
    \begin{equation}
        \begin{aligned}
           & W(\mathcal{T}(\cdot|s_1,k,z_k),\mathcal{T}(\cdot|s_2,k,z_k)) \leq L^\mathcal{S}_\mathcal{T}d_\mathcal{S}(s_1,s_2) \\
           & W(\mathcal{T}(\cdot|s,k_1,z_k),\mathcal{T}(\cdot|s,k_2,z_k)) \leq L^\mathcal{Z}_\mathcal{T}d_\mathcal{K}(k_1,k_2) \\
           & W(\mathcal{T}(\cdot|s,k,z_k),\mathcal{T}(\cdot|s,k,z'_k)) \leq L^\mathcal{K}_\mathcal{T}d_\mathcal{Z}(z_k,z'_k),
        \end{aligned}
    \end{equation}
    where $W$ denotes the Wasserstein Metric. Here, Kronecker delta function are employed to measure the distance: $d_\mathcal{K} (k_1 , k_2) = 1, \forall i \neq j$. In the same way, we can define $(L^\mathcal{S}_\mathcal{R},L^\mathcal{K}_\mathcal{R},L^\mathcal{Z}_\mathcal{R})$-Lipschitz continuous.
\end{definition}

\subsection{Model Predictive Control (MPC)}
MPC is a general framework for model-based control that optimizes action sequences $a_{ t:t+H }$ of finite length such that return is maximized (or cost is minimized) over the time horizon $H$, which corresponds to solving the following optimization problem:
\begin{equation*}
    \pi(s_t) = \underset{a_{t:t+H}}{\text{arg\,max}}\mathbb{E}_\tau\left[ \sum_{i=0}^H \gamma^i \mathcal{R}(s_t,a_t) \right],
\end{equation*}
the return of a candidate trajectory is estimated by simulating it with the learned model \cite{negenborn2005learning}. The agent will use the learned dynamics model to input both the actions and the initial states, obtaining the predicted reward at each time step. The cumulative reward for each action sequence is then computed, and the action sequence with the highest estimated return is selected for execution in the real environment. Various researches \cite{hansen2023td,rubinstein1997optimization,wu2022plan,mazzaglia2022choreographer,eysenbach2022mismatched,amos2018differentiable} have been conducted combining MPC with other training methods to broaden the application domains of model learning.

\section{Learn A Loose yet Flexible Model} \label{Section 3}
In this section, we first show the advantage of Wasserstein Metric over other distribution metrics for dynamics \hyperref[sec 3.A]{(Section 3.A)}. Then, we reveal the relationship between the loss function we carefully designed and the Wasserstein Metric of learned dynamics and environment dynamics, and give the guiding significance of the loss function to dynamics learning \hyperref[sec 3.B]{(Section 3.B)}. Last, we introduce how to optimize the given loss function in practice \hyperref[sec 3.C]{(Section 3.C)}.

\subsection{The Advantage of Wasserstein Metric}\label{sec 3.A}
\begin{figure}
    \centering
    \includegraphics[width=1\linewidth]{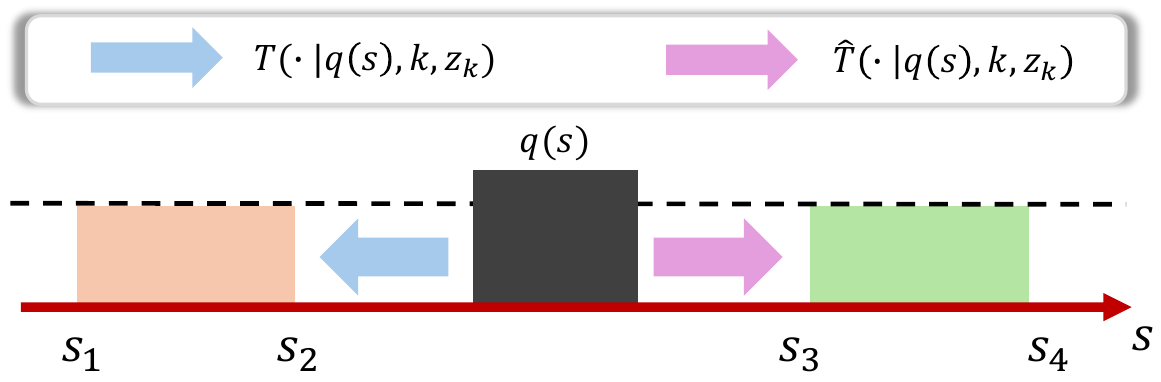}\label{fig1}
    \caption{For simplicity, we designate the state as \textit{scalar}, and the rectangular block represents a distribution. The \textcolor{orange}{orange} rectangle on the left is given by environmental dynamics, while the \textcolor{green}{green} rectangle on the right is given by the learned dynamics, and both rectangles are \textit{congruent}.}
    \label{fig:enter-label}
\end{figure}
\setlength{\belowcaptionskip}{-1cm}
The first concern is why we choose the Wasserstein Metric as the metric for distribution comparison, rather than Kullback-Leibler (KL) divergence \cite{kullback1951information} or Total Variation (TV) distance \cite{chung1989measures}. Based on the definition of the Lipschitz model class \cite{asadi2018lipschitz}, we can define the transition function by accounting for the distribution of states $q(s)$: 
\begin{equation}
    \mathcal{T}(\cdot|q(s),k,z_k) = \int_\mathcal{S} \mathcal{T}(\cdot|s,k,z_k) q(s) ds.
\end{equation}
For simplicity, we assume $q(s)$ follows a uniform distribution across the state space $\mathcal{S}$ and a fixed hybrid action $(k,z_k)$. Moreover, the state distribution after the transition still follows uniform distribution. As illustrated in \hyperref[fig1]{Fig .1}, notably, the state distributions resulting from environment dynamics $\mathcal{T}$ and learned dynamics $\hat{\mathcal{T}}$, respectively, are non-overlapping. In this case, we derive the corresponding KL divergence and TV distance for each:
\begin{equation}
    \begin{aligned}
        & D_\text{KL}(\mathcal{T},\hat{\mathcal{T}}) \\
        & := \int \mathcal{T}(s'|q(s),k,z_k) \text{log} \frac{\mathcal{T}(s'|q(s),k,z_k)}{\hat{\mathcal{T}}(s'|q(s),k,z_k)} ds' = \infty,
    \end{aligned}
\end{equation}
\begin{equation}
     \begin{aligned}
        & D_\text{TV}(\mathcal{T},\hat{\mathcal{T}}) \\
        & := \frac{1}{2} \int \Big| \mathcal{T}(s'|q(s),k,z_k) -  \hat{\mathcal{T}}(s'|q(s),k,z_k)\Big| ds' = 1.
    \end{aligned}
\end{equation}
This reveals that when the KL divergence or TV distance is used as a metric between dynamics, if the environment dynamics and the learned dynamics are non-overlapping, the result will always be a fixed value (or non-existent), leading to indiscriminability and misleading further learning by the agent. In contrast, the Wasserstein Metric avoids the aforementioned issues due to its inherent properties:
\begin{equation}
\begin{aligned}
    & W(\mathcal{T},\hat{\mathcal{T}}) \\
    & := \text{inf}_{\gamma \in \Gamma(\mathcal{T},\hat{\mathcal{T}})} \mathbb{E}_{(v,v') \sim \gamma} [d(v,v')] = \frac{|s_1 + s_2 - s_3 - s_4|}{2},
\end{aligned}
\end{equation}
where $\Gamma(p,q)$ represents the set of all possible joint distributions that can be formed between $\mathcal{T}$ and $\hat{\mathcal{T}}$.

\subsection{Loss Function}\label{sec 3.B}
In many previous MBRL works, two primary types of learning dynamics loss functions are commonly used. The first is the L2 norm of the difference between the model-predicted state and the real state from environment \cite{hamed2024dr,janner2019trust,clavera2018model}, and the second is the value-aware method, which does not require every model-predicted state to be close to the real state but instead ensures that the values between the model and the environment are aligned \cite{vemula2023virtues,farahmand2018iterative,feinberg2018model}. Although these designs have yielded notable results, they also exhibit certain limitations. First, minimizing the L2 norm between learned dynamics and environment dynamics may learn an overly precise model, undermining the exploration capabilities of the agent.\cite{young2022benefits, palenicek2023diminishing, babaeizadeh2020models, lee2023dreamsmooth}. Second, a value-aware approach focuses on value function alignment, which may overlook state information that, although it does not directly influence short-term value, is essential for long-term strategic planning. For example, slight deviations from certain states may not lead to obvious changes in value immediately, but can accumulate, resulting in policy biases in the long-term decision-making process. 

We aim to learn a \textit{loose yet flexible} dynamics model that, on the one hand, does not overly resemble the environment dynamics, thereby inhibiting exploration, while still considering alignment issues throughout the entire rollout horizon. Under Lipstchiz-PAMDPs, we design the following loss function:
\begin{equation} \label{eq7}
    \mathcal{L}^\text{ex}(\tau;f) = \underset{f:K_{d_\mathbb{R},d_\mathbb{R}} \leq 1}{\text{max}}  \sum_{t = t_0}^{t_0 + H} \gamma^{t - t_0} | \mathbb{E}_{s_{t+1}}f(s_{t+1}) - \mathbb{E}_{\tau}f(\hat{s}_{t+1}) |, 
\end{equation}
where $\tau = \{ s_{t_0},k_{t_0},z_{k_{t_0}}, \dots, s_{t_0 + H},k_{t_0 + H},z_{k_{t_0 + H}} \}$ is the $H$-steps sequence sampled from environment, and $\hat{s}_{t+1} \sim \hat{\mathcal{T}}(\cdot|s_t,k_t,z_{k_t})$ is the predicted state obtained by learned dynamics. By employing the equivalent deformation of the Wasserstein Metric, we can derive an upper bound for the loss function as follows:
\begin{theorem}[Looseness] \label{theorem 3.1}
    \textit{Given a $H$-steps sequence $\tau = \{ s_{t_0},k_{t_0},z_{k_{t_0}}, \dots, s_{t_0 + H},k_{t_0 + H},z_{k_{t_0 + H}} \}$ sampled from environment, an upper bound of the loss function $\mathcal{L}^\text{ex}(\tau;f)$ is the $\gamma$-weighted sum for the Wasserstein Metric between environment dynamics $\mathcal{T}$ and learned dynamics $\hat{\mathcal{T}}$ at each timestep:}
    \begin{equation}
        \mathcal{L}^\text{ex}(\tau;f) \leq \sum_{t=t_0}^{t_0 + H} \gamma^{t-t_0} W(\mathcal{T}(\cdot|s_t,k_t,z_{k_t}),\hat{\mathcal{T}}(\cdot|s_t,k_t,z_{k_t})).
    \end{equation}
\end{theorem}
\begin{IEEEproof}
    See \hyperref[Appendix B]{Appendix B}.
\end{IEEEproof}
~\\
\hyperref[theorem 3.1]{Theorem 3.1} proposes that the loss function is \textit{loose}. In the case of the Wasserstein Metric, we can deem that this upper bound represents the cumulative sum of the $\gamma$-weighted errors between the true and predicted states at each time step. Thus, minimizing the loss function $\mathcal{L}^\text{ex}$ without aggressively reducing the cumulative sum of the $\gamma$-weighted errors prevents the learning of an overly precise dynamics model, which distinguishes it from previous approaches.

However, an additional issue arises: while we have established the looseness of the loss function, how can we demonstrate that it facilitates effective learning of dynamics? Therefore, we present the following theorem:
\begin{theorem}[Flexibility]\label{Theorem 3.2}
\textit{Given the same conditions as \hyperref[theorem 3.1]{Theorem 3.1}, an lower bound of the loss function $\mathcal{L}^\text{ex}(\tau;f)$ is the Wasserstein Metric between environment dynamics $\mathcal{T}$ and learned dynamics $\hat{\mathcal{T}}$ at each timestep:}
\begin{equation}
    \begin{aligned}
        & \mathcal{L}^\text{ex}(\tau;f) \geq W(\mathcal{T}          
          (\cdot|s_t,k_t,z_{k_t}),\hat{\mathcal{T}}(\cdot|s_t,k_t,z_{k_t})) \\
        & \forall t \in \{t_0,\dots,t_0 + H  \}.
    \end{aligned}
\end{equation}    
\end{theorem}
\begin{IEEEproof}
    See \hyperref[Appendix C]{Appendix C}.
\end{IEEEproof}
~\\
\hyperlink{theorem 3.2}{Theorem 3.2} indicates that when the loss function is minimized, at least the Wasserstein Metric between the environment dynamics and the learned dynamics can be reduced at \textit{a given time step}, that is, just one-step. Notably, \hyperlink{theorem 3.2}{Theorem 3.2} focuses solely on a single state within the sample trajectory $\tau$. In the context of a continuous state space, our primary interest lies in the Wasserstein Metric derived from the state distribution, as follows:
\begin{corollary}[State distribution]\label{Corollary 3.1}
    \textit{Given the initial state distribution $q(s_t)$, the Wasserstein Metric between the corresponding environment dynamics $\mathcal{T}(\cdot|q(s_t),k_t,z_{k_t})$ and the learned dynamics $\hat{\mathcal{T}}(\cdot|q(s_t),k_t,z_{k_t})$ serves as a lower bound for the loss function:}
    \begin{equation}
        \mathcal{L}^\text{ex}(\tau;f) \geq W(\mathcal{T}          
          (\cdot|q(s_t),k_t,z_{k_t}),\hat{\mathcal{T}}(\cdot|q(s_t),k_t,z_{k_t})).
    \end{equation}
\end{corollary}
\begin{IEEEproof}
\begin{equation}
    \begin{aligned}
        & W(\mathcal{T}(\cdot|q(s_t),k_t,z_{k_t}),\hat{\mathcal{T}} 
          (\cdot|q(s_t),k_t,z_{k_t})) \\
        & = \underset{f}{\text{sup}} \int \int (\hat{\mathcal{T}} (s_{t+1}|s_t,k_t,z_{k_t}) \\
        & \quad \quad - \mathcal{T}(s_{t+1}|s_t,k_t,z_{k_t})) f(s_{t+1})q(s_t) ds_tds_{t+1} \\
        & \leq \int \underset{f}{\text{sup}} \int (\hat{\mathcal{T}} (s_{t+1}|s_t,k_t,z_{k_t}) - \mathcal{T}(s_{t+1}|s_t,k_t,z_{k_t})) \\
        & \quad \quad f(s_{t+1})ds_{t+1} q(s_t)ds_t \\
        & = \int W(\mathcal{T}(\cdot|s_t,k_t,z_{k_t}),\hat{\mathcal{T}}(\cdot|s_t,k_t,z_{k_t})) q(s_t) ds_t \\
        & \leq \int \mathcal{L}^\text{ex}(\tau;f) q(s_t) ds_t \quad \quad \quad \textbf{Theorem 3.2} \\
        & = \mathcal{L}^\text{ex}(\tau;f).
    \end{aligned}
\end{equation}
For line 1 to line 2, we employ the Kantorovich-Rubinstein duality \cite{villani2009optimal} to perform equivalent transformations for Wasserstein Metric. Notably, $f$ needs to meet the Lipschitz condition.
\end{IEEEproof}
~\\
We can use \hyperref[Corollary 3.1]{Corollary 3.1} to learn a flexible dynamics on a continuous state space. However, it should be noted that the loss function we propose is learned in the range of horizon $H$, thus we also need to investigate its influence on dynamics learning with horizon $H$. The following corollary provides the answer:
\begin{corollary}($H$-steps error)\label{Corollary 3.2}
    \textit{We denote the transition function of state $s_t$ transfering to $s_{t+H}$ as $\mathcal{T}^H(\cdot|q(s_t),k_t,z_{k_t})$, where $s_t \sim q(s_t)$. Given the same condition as \hyperref[theorem 3.1]{Theorem 3.1}, we can derive that}
    \begin{equation}
         \mathcal{L}^\text{ex}(\tau;f) \geq \frac{W(\mathcal{T}^H(\cdot|q(s_t),k_t,z_{k_t}),\hat{\mathcal{T}}^H 
          (\cdot|q(s_t),k_t,z_{k_t}))}{\sum_{i=0}^{H-1}\left(L_{\overline{\mathcal{T}}}^{\mathcal{S}}\right)^i},
    \end{equation}
\textit{where $L_{\overline{\mathcal{T}}}^{\mathcal{S}} = {\rm min}\left( L_{\mathcal{T}}^{\mathcal{S}}, L_{\hat{\mathcal{T}}}^{\mathcal{S}}\right)$, $L_{\hat{\mathcal{T}}}^{\mathcal{S}}$ is the Lipschitz constant with state as variable in learned dynamics.}
\end{corollary}
\begin{IEEEproof}
\begin{equation}
    \begin{aligned}
        & W(\mathcal{T}^H(\cdot|q(s_t),k_t,z_{k_t}),\hat{\mathcal{T}}^H 
          (\cdot|q(s_t),k_t,z_{k_t})) \\
        & = W\Big(\mathcal{T}\Big(\mathcal{T}^{H-1}(\cdot|q(s_t),k_t,z_{k_t})\Big),\hat{\mathcal{T}}\Big(\hat{\mathcal{T}}^{H-1}(\cdot|q(s_t),k_t,z_{k_t})\Big)\Big) \\
        & \leq \mathcal{L}^\text{ex}(\tau;f) + L_{\overline{\mathcal{T}}}^{\mathcal{S}} W(\mathcal{T}^{H-1}(\cdot|q(s_t),k_t,z_{k_t}),\hat{\mathcal{T}}^{H-1}(\cdot|q(s_t),k_t,z_{k_t}))\\
        & \dots \\
        & \leq \mathcal{L}^\text{ex}(\tau;f) \sum_{i=0}^{H-1}\left(L_{\overline{\mathcal{T}}}^{\mathcal{S}}\right)^i
          (\cdot|q(s_t),k_t,z_{k_t})),
    \end{aligned}
\end{equation}
we employ Composition Lemma \cite{asadi2018lipschitz} to make the conversion from line 2 to line 3.
\end{IEEEproof}
~\\
\hyperref[Corollary 3.2]{Corollary 3.2} suggests the potential to minimize the loss function, thereby indirectly reducing the dynamics error over $H$-steps. However, the denominator of this lower bound is the \textit{state-dependent Lipschitz constant} of the transition function, which is further related to the learned dynamics as defined by Lipschitz-PAMDPs. Therefore, to effectively reduce the dynamics error over $H$-steps, we also need to control the Lipschitz constant.

\subsection{Practical Implementation}\label{sec 3.C}
In practice, we utilize a neural network (NN) to fit $f$. We denote the weight parameter of this NN as $\mathbf{W}$, while the corresponding parameter of learned dynamics as $\phi$. Here, we give the loss function used in the practical implementation:
\begin{equation}
    \begin{aligned}
        & \mathcal{L}^\text{ex}(\tau;\mathbf{W},\phi) \\
        & = \underset{f_{\mathbf{W}}:K_{d_\mathbb{R},d_\mathbb{R}} \leq 1}{\text{max}}  \sum_{t = t_0}^{t_0 + H} \gamma^{t - t_0} | \mathbb{E}_{s_{t+1}}f(s_{t+1}) - \mathbb{E}_{\tau}f(\hat{s}_{t+1}) |.
    \end{aligned}
\end{equation}
Notably, although the learned dynamics do not explicitly appear in the loss function, the predicted state $\hat{s}_{t+1}$ is derived from learned dynamics, that is, $\hat{s}_{t+1} \sim \mathcal{T}_\phi(\cdot|s_t,k_t,z_{k_t})$, making the loss function dependent on $\phi$ as a variable. The approach to minimizing this loss function is akin to the principles of \cite{goodfellow2014generative,arjovsky2017towards,arjovsky2017wasserstein} , specifically by maximizing the value of function $f$ (acting as a discriminator) while simultaneously minimizing the discrepancy between the environment dynamics and the learned dynamics (analogous to a generator). However, there are three remaining problems: (1) How to obtain the expectation of $f(s_{t+1})$ and $f(\hat{s}_{t+1})$? (2) How to constrain the Lipschitz constant on $f$? (3) How to accomplish the optimizations revealed by \hyperref[Corollary 3.2]{Corollary 3.2}? 

Based on the characteristics of model roll-out, we can easily complete a process similar to beam search \cite{lowerre1976harpy,wu2016google,wang2024walk}, collect true state and predicted state in multiple roll-out processes, so as to obtain the expectation. As for the Lipstchiz constraint on $f$, we can accomplish it by introducing spectral regularization \cite{yoshida2017spectral} (the derivation can be refered to \hyperref[Appendix D]{Appendix D}), that is, minimizing $(|| \textbf{W} ||_2 - 1)^2$. For the last question, one intuition is that if environment dynamics and learned dynamics are close enough, then $\mathcal{L}^{\mathcal{S}}_{\overline{\mathcal{T}}}$ is almost fixed. This is evident because the environment dynamics are governed solely by the environment. Once the environment is determined, the Lipschitz constant of its transition function is fixed. Thus, we aim to implement a more strict loss optimization to further constrain the distance between the environment dynamics and the learned dynamics:
\begin{equation} \label{eq15}
    \begin{aligned}
        & \mathcal{L}^{\text{total}}_{\text{dyn}} = \mathcal{L}^{\text{org}}_{\text{dyn}} + \lambda \mathcal{L}^{\text{ex}}, \\
        & \text{where} \ \mathcal{L}^{\text{org}}_{\text{dyn}} = \mathbb{E}_\tau \sum_{t=t_0}^{t_0 + H} \gamma^{t-t_0} || s_{t+1} - \hat{s}_{t+1} ||^2_2.
    \end{aligned}
\end{equation}
This means that we control how flexible the learned dynamics are by adjusting the coefficient $\lambda$.

\section{Reward Smoothing in Stable Training Phase} \label{Section 4}
\begin{figure}[t]
    \centering
    \includegraphics[width=1\linewidth]{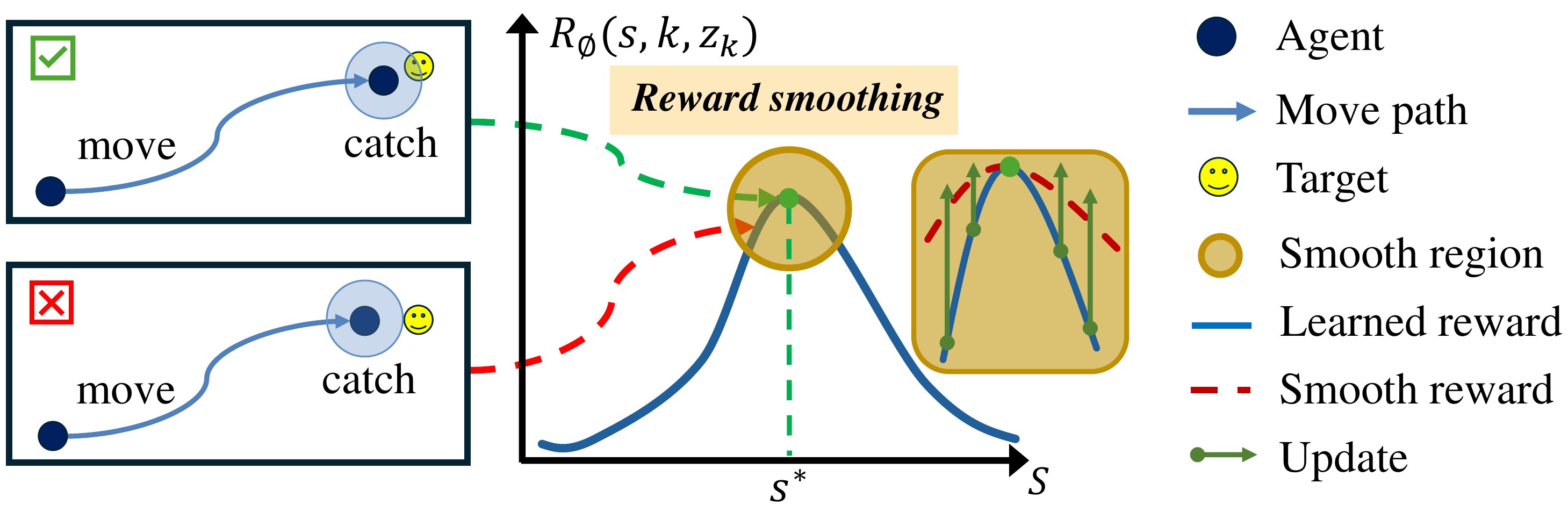} \label{Fig. 2}
    \caption{An intuitional example of reward smoothing for \texttt{Catch Point} environment. In \texttt{Catch Point}, the agent can either move or execute a catch, with the latter having a specific range indicated by the light \textcolor{blue}{blue} area. Upon successfully capturing the target, the agent receives the highest reward, which corresponds to the maximum value of the reward function. If the agent is merely close to capturing the target, it will not receive a high reward, as the conditions for successful capture have not been met. Given the continuity of the state space, the corresponding state will reside within a small neighborhood of the optimal state, as indicated by the \textcolor{yellow}{yellow} region, which represents the region where we aim to implement reward smoothing.}
    \label{fig:enter-label}
\end{figure}
In previous MBRL algorithms, particular emphasis should be placed on accurately learning the reward function. An intuitive approach suggests that the reward function must be learned with sufficient precision to accurately reflect the policy's effectiveness in real-world scenarios, particularly when we execute the model rollouts. However, recent researches\cite{babaeizadeh2020models,lee2023dreamsmooth,young2022benefits} indicate that there is no strong correlation between reward function error and asymptotic performance. In fact, \textit{within a specific range of error, larger reward prediction errors may even lead to better agent performance.} This phenomenon could result from overfitting to observed states, coupled with the agent’s pessimistic exploration, which directs the policy toward familiar states with high predicted rewards, rather than encouraging exploration of novel states that could offer higher rewards. As a result, a random policy gradually collapse into deterministic one, preventing the agent from escaping local minima. 

Based on this justification, we design reward smoothing for FLEXplore, which enables the agent to pay attention to other states in the neighborhood of high reward states and avoid pessimistic exploration by agents. \hyperref[Fig. 2]{Fig. 2} shows an intuitional example of \texttt{Catch Point} \cite{li2021hyar}. In practice, it is not feasible to smooth every point within a smooth region, instead, we generate certain perturbed states $\tilde{s}$ for smoothing:
\begin{equation} \label{eq16}
\begin{aligned}
    & \mathcal{L}^{\text{smt}}(\epsilon,\tau;\psi) \\
    & = \sum_{t=t_0}^{t_0+H} \underset{|| s_t - \tilde{s}_t ||_\infty \leq \epsilon}{\text{max}} \big( \mathcal{R}_\psi(s_t,k_t,z_{k_t}) - \mathcal{R}_\psi(\tilde{s}_t,k_t,z_{k_t}) \big)^2.
\end{aligned}
\end{equation}
We employ Fast Gradient Sign Method (FGSM) \cite{goodfellow2014explaining} to generate the perturbed states:
\begin{equation}
    \begin{aligned} \label{eq17}
        & \tilde{s} = s' + \epsilon\,\text{sign}\big( \nabla_{\tilde{s}|\tilde{s} = 
          s'} (\mathcal{R}_\psi(s,k,z_{k}) - \mathcal{R}_\psi(\tilde{s},k,z_{k}))^2 \big), \\
        & \text{where} \ s' \sim \mathcal{N}(s,\epsilon^2\mathbf{I}).
    \end{aligned}
\end{equation}
It is essential to emphasize that we do not apply FGSM directly to the original state $s$, as the theoretical derivative of $s$ in the original state is zero. Consequently, we utilize a multivariate Gaussian distribution to generate a slight offset $s'$ around $s$, after which we perform the FGSM operation. 

In practice, to prevent excessive modification of the learned reward function, we apply smoothing exclusively to the maximum reward at each time step. Notably, we execute reward smoothing in the stationary training phase. In the early training phase, the discrepancy between the environment reward function and the learned reward function is not acceptable. Executing reward smoothing too early will lead to instability in model learning. Therefore, for the learning of the reward function, the corresponding loss function is
\begin{equation} \label{eq18}
\begin{aligned}
    & \mathcal{L}^{\text{total}}_{\text{rew}} = \left\{
    \begin{aligned}
        & \mathcal{L}_{\text{rew}}^{\text{org}}, \quad &\text{if} \ timestep \leq T \\
        & \mathcal{L}_{\text{rew}}^{\text{org}} + \mu\mathcal{L}^{\text{smt}}, \quad &\text{if} \ timestep > T \\
    \end{aligned}
    \right. \\
    & \text{where} \ \mathcal{L}_{\text{rew}}^{\text{org}} = \mathbb{E}_\tau \sum_{t=t_0}^{t_0 + H} \gamma^{t-t_0} || r_t - \hat{r}_{t} ||^2_2,
\end{aligned}
\end{equation}
where $T$ is the reward smoothing threshold for total model learning timestep.

\begin{figure*}[t]
    \centering
    \includegraphics[width=1\linewidth]{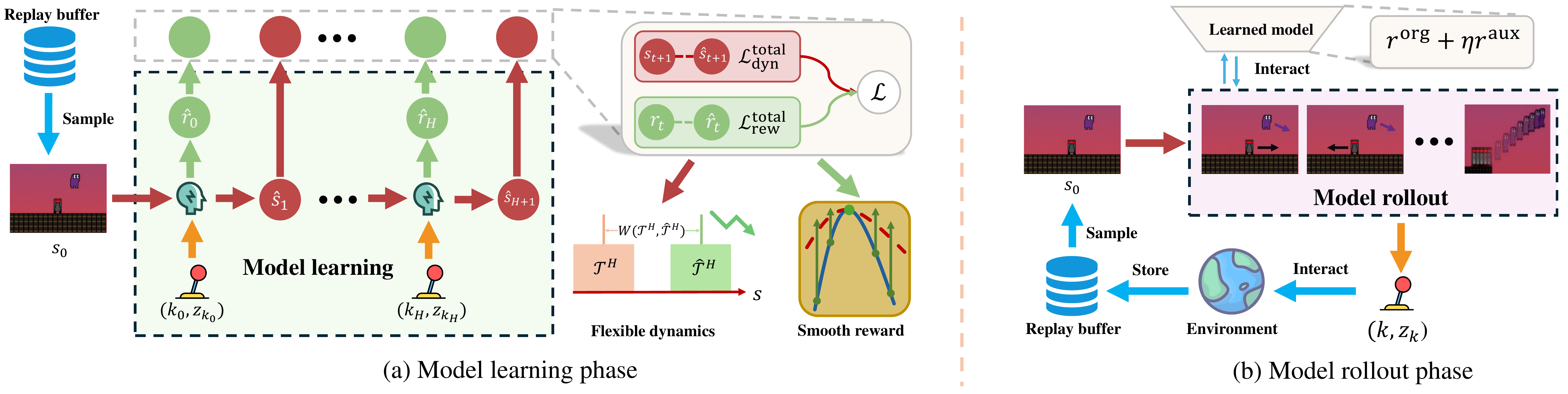} \label{Fig. 3}
    \caption{The framework of \textbf{FLEXplore}. In the model learning phase (the left subplot), the agent samples sequence $\tau = \{ s_0,k_0,z_{k_0},k_1,z_{k_1},\dots,k_H,z_{k_H} \}$ from the replay buffer and then employs it within the model to generate predicted states $\hat{s}_{t+1}$ and rewards $\hat{r}_t$ for each timestep $t$. It is trained with $\mathcal{L}^{\text{total}}_{\text{dyn}}$ and $\mathcal{L}^{\text{total}}_{\text{rew}}$ respectively, which correspond to flexible model learning in \hyperref[Section 3]{Section 3} and reward smoothing in \hyperref[Section 4]{Section 4}. In the model rollout phase (the right subplot), the agent interacts using a hybrid action sequence on the learned model. In particular, we maximize mutual information as an auxiliary reward signal $r^{\text{aux}}$ to enhance agent exploration. The agent then interacts with the environment by selecting the first hybrid action $(k,z_k)$ from the hybrid action sequence with the highest trajectory return and stores the resulting feedback in the replay buffer.}
    \label{fig:enter-label}
\end{figure*}

\section{Encourage Exploration by Maximizing Mutual Information}
In certain highly complex environments, the number of continuous parameters increases exponentially as the number of discrete actions grows. For example, in \texttt{Hard Move} \cite{li2021hyar}, if the number of discrete actions is $n$, the corresponding number of continuous parameters becomes an overwhelming $2^n$. Even in the absence of sparse reward issues, learning policies through negative reward signals remains challenging or almost impossible when interacting with the environment in such an extensive action space. Li et al. \cite{li2021hyar} points out that no regular model-free PAMDPs algorithms can learn a meaningful policy without learning a latent action embedding space. However, the unavoidable loss of information during the transition from the raw space to the latent space poses a significant limitation for HyAR \cite{li2021hyar}. In response, we have introduced two key improvements. First, we employ the model-based PAMDPs algorithm, and the performance of DLPA \cite{zhang2024model} demonstrates the feasibility of applying model-based learning policies directly on the raw space. Second, we leverage mutual information as an auxiliary signal to incentivize effective exploration by the agents, thereby enhancing the efficiency of policy learning. In particular, we maximize the mutual information of the next timestep state $s'$ and the current timestep hybrid action $(k,z_k)$:
\begin{equation*}
    \underset{\Phi}{\text{max}} \, I(s';(k,z_k)|s,\Phi),
\end{equation*}
where $s$ is the current timestep state, while $\Phi$ is the set of parameters related to the whole process. However, it is difficult to optimize this mutual information directly, so we give the following proposition for indirect optimization:
\begin{proposition}[Variational lower bound]\label{Proposition 5.1}
    \textit{Given the conditional continuous parameter distribution $p_{\theta}(z_k|s,k)$ and learned dynamics $\mathcal{T}_{\phi}(s'|s,k,z_k)$, a variational lower bound of mutual information between $s'$ and $(k,z_k)$ is:}
    \begin{equation}
        I(s';(k,z_k)|s,\Phi) \geq \mathbb{E}[{\rm log} \,\mathcal{T}_{\phi}(s'|s,k,z_k) - {\rm log} \,p_{\theta}(z_k|s,k)],
    \end{equation}
\end{proposition}
\begin{IEEEproof}
   We set $\mathcal{H}$ as the notation for entropy. Using the chain rule, non-negativity, and symmetry of mutual information respectively, we can derive that
    \begin{equation}
        \begin{aligned}
            I(s';(k,z_k)|s,\Phi) 
            &= I(s';k|s,\Phi) + I(s';z_k|k,s,\Phi) \\
                & \geq I(s';z_k|k,s,\Phi) \quad \\
                & = I(z_k;s'|k,s,\Phi) \quad \\
                & = \mathcal{H}(z_k|s,k,\theta) - \mathcal{H}(s'|s,k,z_k,\phi).
        \end{aligned}
    \end{equation}
We employ a variational lower-bound approximation \cite{barber2004algorithm} as follows:
\begin{equation}
    \begin{aligned}
        & \mathcal{H}(z_k|s,k,\theta) - \mathcal{H}(s'|s,k,z_k,\phi) \\
        & \geq \mathbb{E}[{\rm log} \,\mathcal{T}_{\phi}(s'|s,k,z_k) - {\rm log} \,p_{\theta}(z_k|s,k)].
    \end{aligned}
\end{equation}
\end{IEEEproof}
~\\
\hyperref[Proposition 5.1]{Proposition 5.1} gives a viable method for indirectly maximizing target mutual information by enhancing the certainty of learned dynamics while introducing controlled randomness into continuous parameter networks. To incorporate randomness into continuous parameters, we model them using multivariate Gaussian:
\begin{equation}
    z_k \sim \mathcal{N}(\mu_k,\sigma_k^2 \mathbf{I}), \quad \mu_k,\sigma_k \sim \mathbb{R}^{|z_k|}.
\end{equation}
This implies that each continuous parameter is conditioned on its corresponding discrete action, with both the mean and variance deciding the parameters. In practice, as the selection of discrete actions is stochastic (typically modeled with a categorical distribution), we aim to enable gradient flow through this process, and thus employ Gumbel-Softmax \cite{jang2016categorical} in place of the categorical distribution.

We optimize mutual information throughout the model’s rollout phase by defining auxiliary reward signals:
\begin{equation} \label{eq23}
    \begin{aligned}
        & r^\text{total} = r^\text{org} + \eta r^\text{aux}, \\
        & \text{where} \ r^\text{aux} = \mathbb{E}[{\rm log} \,\mathcal{T}_{\phi}(s'|s,k,z_k) - {\rm log} \,p_{\theta}(z_k|s,k)].
    \end{aligned}
\end{equation}
Notably, \textit{we do not use this auxiliary reward signal during the learning phase of the model}, but during model roll-out with the learned model. 

\section{How FLEXplore Reduces the Regret of the Rollout Trajectory?}
\hyperref[Fig. 3]{Fig. 3} shows the framework of FLEXplore. However, a core problem in FLEXplore is \textit{how to use the model to learn an optimal policy that results in an optimal trajectory}, and the quality of the policy is typically evaluated through the regret of the rollout trajectory $\hat{\zeta} = \{ \hat{s}_t,\hat{k}_t,\hat{z}_{k_t} \}_{1:H}$ \cite{hazan2016introduction,zhang2024model}:
\begin{equation}
    | \mathcal{J}_{\zeta^*} - \mathcal{J}_{\hat{\zeta}} | := \sum_{t=1}^H \gamma^{t-1} | \mathcal{R}(s_t,k_t,z_{k_t}) - \mathcal{R}_\psi(\hat{s}_t,\hat{k}_t,\hat{z}_{k_t}) |, 
\end{equation}
where $z_{k_t} \sim p(\cdot|k_t)$, $\hat{z}_{k_t} \sim p_{\theta}(\cdot|\hat{k}_t)$. We will next present the regret upper bound theorem, which demonstrates how FLEXplore minimizes regret and enhances asymptotic performance:
\begin{theorem}[the regret upper bound, proved by \cite{zhang2024model}] \label{Theorem 6.1}
    \textit{Given a Lipschitz-PAMDP, the learned reward function $\mathcal{R}_\psi$ is $\epsilon_{\mathcal{R}}$-accurate, that is, $|\mathcal{R}(s,k,z_k) - \mathcal{R}_\psi(s,k,z_k)| \leq \epsilon_{\mathcal{R}}$. If $L^{\mathcal{S}}_{\overline{\mathcal{T}}} < 1$, then the regret of the rollout trajectory $\hat{\zeta}$ is bounded by:}
    \begin{equation}
        \begin{aligned}
            & | \mathcal{J}_{\zeta^*} - \mathcal{J}_{\hat{\zeta}} | \\
            & \leq \mathcal{O}\Big(m(L^{\mathcal{K}}_{\overline{\mathcal{R}}} + \textcolor{red}{L^{\mathcal{S}}_{\overline{\mathcal{R}}}}L^{\mathcal{K}}_{\overline{\mathcal{T}}} \\
            & + H(\epsilon_{\mathcal{R}} + \textcolor{red}{L^{\mathcal{S}}_{\overline{\mathcal{R}}}}\textcolor{orange}{\mathcal{L}^\text{ex}} + (L^{\mathcal{Z}}_{\overline{\mathcal{R}}}) + \textcolor{red}{L^{\mathcal{S}}_{\overline{\mathcal{R}}}}L^{\mathcal{Z}}_{\overline{\mathcal{T}}})(\frac{m}{H} \Delta_{k,\hat{k}} + \Delta_{p,p_\theta})\Big),
        \end{aligned}
    \end{equation}
    \textit{where $m = \sum_{t=1}^H \mathbb{I}\{ k_t \neq \hat{k}_t \}$, $\Delta_{k,\hat{k}} = W(p(\cdot|k),p(\cdot|\hat{k}))$, $\Delta_{p,p_\theta} = W(p(\cdot|k),p_{\theta}(\cdot|k))$}.
\end{theorem}
\begin{IEEEproof}
    According to \hyperref[Theorem 3.2]{Theorem 3.2}, $\mathcal{L}^\text{ex}$ is an upper bound of $W(\mathcal{T}(s,k,z_k),\mathcal{T}_\phi(s,k,z_k))$. See \cite{zhang2024model} for the remaining proof. 
\end{IEEEproof}
~\\
In \hyperref[Theorem 6.1]{Theorem 6.1}, we highlight the parts of flexible dynamics learning in \textcolor{orange}{orange font} and highlight the parts of reward smoothing in \textcolor{red}{red font}. Obviously, optimizing the loss function $\mathcal{L}^\text{ex}$ we designed can be an effective way to reduce the upper bound. Additionally, the method taken for reward smoothing (\hyperref[eq16]{Equation 16}) can also reduce the upper bound. \cite{zheng2023model} points out that the local Lipschitz constant can be reduced in a manner similar to \hyperref[eq16]{Equation 16}. By \hyperref[def2.1]{Definition 2.1}, reducing the local Lipschitz constant is useful for reducing the global Lipschitz constant, thus deriving
\begin{equation}
    \begin{aligned}
       & L^{\mathcal{S}}_{\hat{\mathcal{R}}_1} \leq L^{\mathcal{S}}_{\hat{\mathcal{R}}_2} \\
       & \Rightarrow L^{\mathcal{S}}_{\overline{\mathcal{R}}_1} = \text{min} \{  L^{\mathcal{S}}_{\hat{\mathcal{R}}_1},  L^{\mathcal{S}}_{\mathcal{R}} \} \leq \text{min} \{  L^{\mathcal{S}}_{\hat{\mathcal{R}}_2},  L^{\mathcal{S}}_{\mathcal{R}} \} = L^{\mathcal{S}}_{\overline{\mathcal{R}}_2}.
    \end{aligned}
\end{equation}
The implementation of FLEXplore is simple. Compared with existing MPC algorithms, FLEXplore only requires only three changes, but it can achieve a better and more amazing result. In \hyperref[Appendix E]{Appendix E}, we provide the pseudocode for FLEXplore.

\begin{figure*}[t] \label{Fig. 4}
    \centering
    \includegraphics[width=0.9\linewidth]{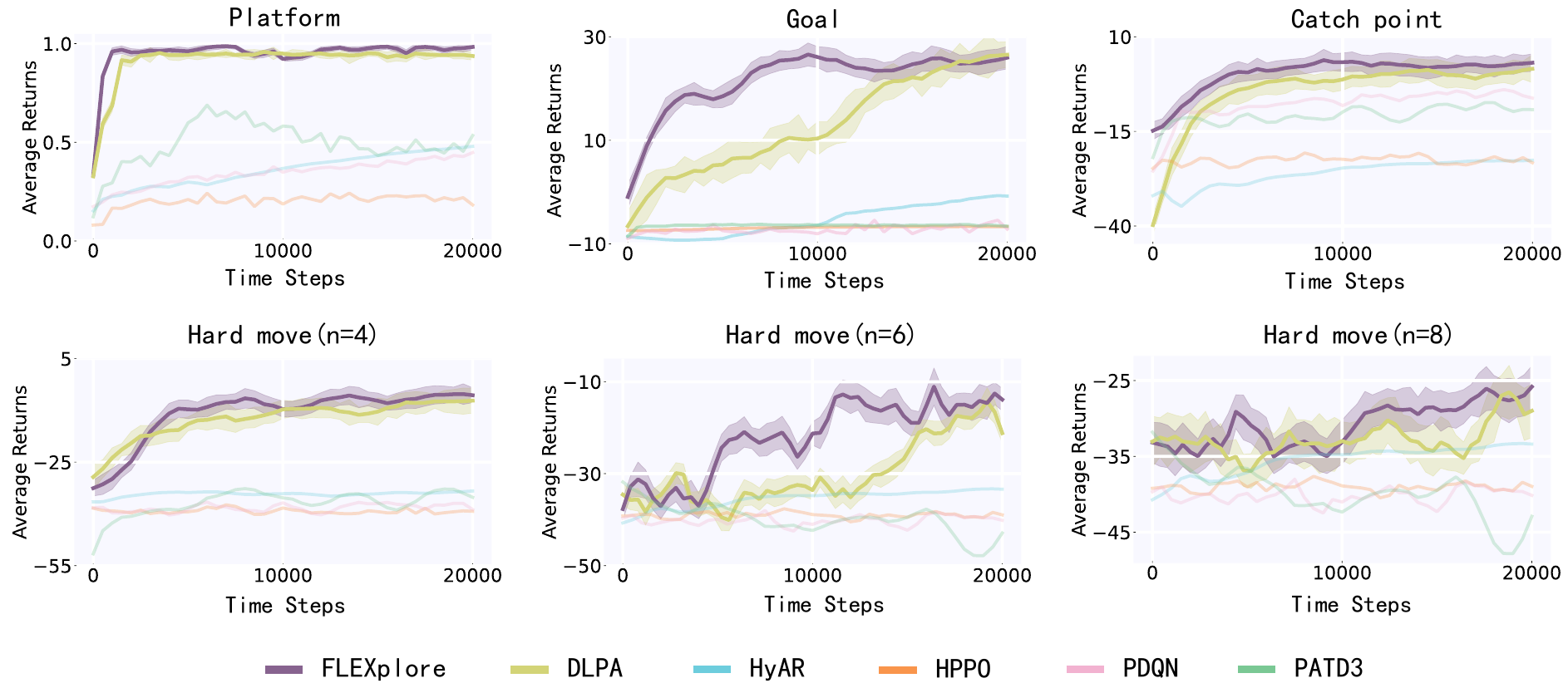}
    \caption{Performance evaluation of FLEXplore and baselines over 4 seeds across six standard PAMDP benchmarks. To demonstrate the agent's learning efficiency, we plot the average return over the first 20,000 time steps. FLEXplore achieves the best performance in the first 20,000 time steps on all the six benchmarks.}
    \label{fig:enter-label}
\end{figure*}
\section{Experiments}
In this section, we test the performance of FLEXplore on six standard PAMDP benchmarks (\hyperref[Appendix F]{Appendix F} gives a brief introduction to these benchmarks) and address the following research questions:
\begin{itemize}
    \item \textbf{RQ1: Does FLEXplore improve learning efficiency and asymptotic performance compared to other baselines?} (\hyperref[Section 7.A]{Section 7.A})
    \item \textbf{RQ2: Does FLEXplore learn a more flexible dynamics compared to DLPA?} (\hyperref[Section 7.B]{Section 7.B})
    \item \textbf{RQ3: Does the introduction of reward smoothing improve FLEXplore's performance?} (\hyperref[Section 7.C]{Section 7.C})
    \item \textbf{RQ4: Does the auxiliary reward signal enhance the agent's exploration?} (\hyperref[Section 7.D]{Section 7.D})
\end{itemize}
In the choice of baselines, for model-based method, we select \textbf{DLPA} \cite{zhang2024model}, which learns a model trained through MPC to enhance the sample effciency. For model-free methods, we select \textbf{HyAR} \cite{li2021hyar}, which learns a compact and decodable latent representation space for the original hybrid action space, \textbf{HPPO} \cite{fan2019hybrid}, which uses multiple policy heads consisting of one for discrete actions and the others for corresponding continuous parameter of each discrete action separately, \textbf{PDQN} \cite{xiong2018parametrized}, which proposes a hybrid structure of DQN and DDPG, and \textbf{PATD3}, which is the variant of PADDPG \cite{Masson2016ParamActions}. 

\subsection{Overall Performance} \label{Section 7.A}
In \hyperref[Fig. 4]{Fig. 4}, we show the averaged return in the early training phase to demonstrate that FLEXplore's outstanding learning efficiency. On all the six benchmarks, FLEXplore achieves the best performance in the first 20,000 time steps. Since model-free algorithms (HyAR, HPPO, PDQN and PATD3) necessitate extensive interactions with the environment to learn effective policies, they often fail to acquire satisfactory policies promptly enough to yield a high trajectory return during the early phases of training. Even in comparison to model-based algorithm (DLPA), FLEXplore has demonstrated superior learning efficiency. Another question is how about the FLEXplore's asymptotic performance. \hyperref[tab1]{Table 1} shows the asymptotic performance of each algorithm, FLEXplore achieves the best performance in four of the benchmarks. This indicates that, compared to the previous PAMDPs algorithms, FLEXplore enhances both learning efficiency and asymptotic performance, aligning with our theoretical analysis.
\begin{table*}[t] \label{tab1}
    \centering
    \caption{The asymptotic performance of different algorithms on all the six benchmarks at the end of training.}
    \setlength{\tabcolsep}{11.7pt}
    \begin{tabular}{l|c|c|c|c|c|c}
        \toprule
        ~ & \textbf{Platform} & \textbf{Goal} & \textbf{Catch point} & \textbf{Hard move(4)} & \textbf{Hard move(6)} & \textbf{Hard move(8)} \\
        \midrule
        \rowcolor{gray!20}
        \textbf{FLEXplore} & $\mathbf{0.97 \pm 0.02}$ & $32.21 \pm 3.07$ & $\mathbf{8.49 \pm 2.13}$ & $\mathbf{6.13 \pm 2.27}$ & $\mathbf{6.97 \pm 3.39}$ & $6.22 \pm 6.10$ \\
        \textbf{DLPA} \cite{zhang2024model} & $0.94 \pm 0.04$ & $29.97 \pm 4.02$ & $7.74 \pm 2.04$ & $5.87 \pm 2.98$ & $6.55 \pm 4.02$ & $\mathbf{7.22 \pm 8.21}$ \\
        \textbf{HyAR} \cite{li2021hyar} & $0.95 \pm 0.04$ & $\mathbf{34.12 \pm 2.99}$ & $5.24 \pm 3.77$ & $6.01 \pm 1.83$ & $6.21 \pm 2.12$ & $0.03 \pm 4.01$ \\
        \textbf{HPPO} \cite{fan2019hybrid} & $0.79 \pm 0.03$ & $-6.76 \pm 0.46$ & $4.22 \pm 2.98$ & $-32.19 \pm 4.29$ & $-32.01 \pm 6.28$ & $-36.54 \pm 8.55$ \\
        \textbf{PDQN} \cite{xiong2018parametrized} & $0.83 \pm 0.05$ & $32.07 \pm 4.22$ & $6.23 \pm 2.21$ & $5.01 \pm 3.98$ & $-19.22 \pm 7.90$ & $-38.20 \pm 4.01$ \\
        \textbf{PATD3} & $0.90 \pm 0.07$ & $-3.91 \pm 8.27$ & $3.88 \pm 10.21$ & $-9.88 \pm 5.29$ & $-38.72 \pm 17.75$ & $-40.13 \pm 16.97$ \\
        \bottomrule
    \end{tabular}
    \label{tab:my_label}
\end{table*}

\begin{figure}[t] \label{Fig. 5}
    \centering
    \includegraphics[width=1\linewidth]{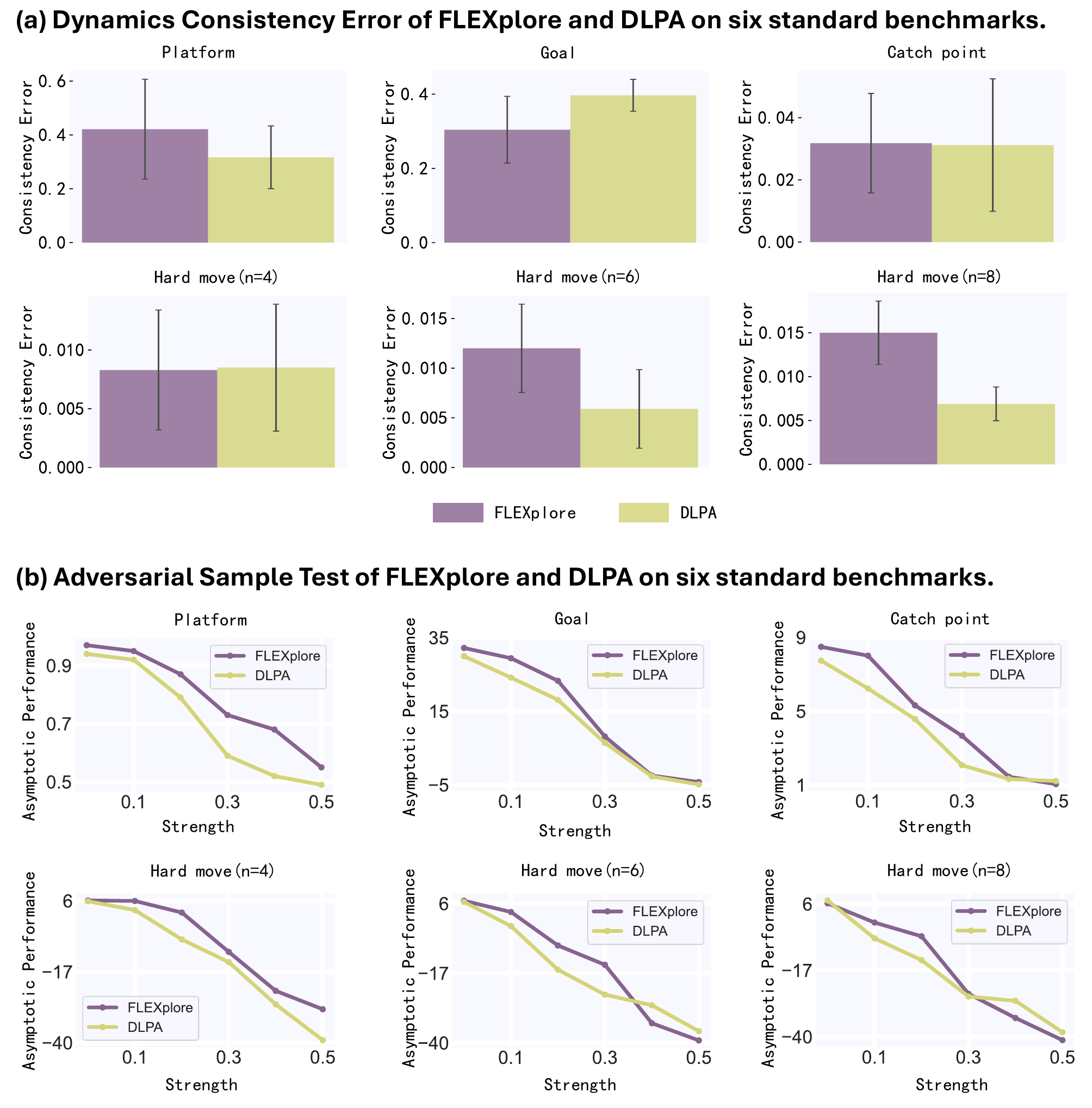}
    \caption{Investigation on loose yet flexible dynamics learning. (a) Dynamics consistency error of FLEXplore and DLPA, which shows FLEXplore can learn a looser dynamics. (b) Asymptotic performance of FLEXplore and DLPA under adversarial sample test with different strengths, which shows FLEXplore can learn a more flexible dynamics.}
    \label{fig:enter-label}
\end{figure}

\subsection{Loose yet Flexible Dynamics Learning} \label{Section 7.B}
In this section, we demonstrate that FLEXplore has learned a loose yet flexible dynamics through the elaborated loss functions.
\subsubsection{Dynamics Consistency Error}
To assess the differences in the looseness of dynamics learned by FLEXplore and DLPA, we introduce the concept of \textit{dynamics consistency error}. In simple terms, the \textit{dynamics consistency error} is defined as the averaged L2 norm of the difference between the real state and the predicted state over the entire training process. Clearly, a larger dynamics consistency error indicates that the learned dynamics are more dissimilar in comparison to the environment dynamics. 

\hyperref[Fig. 5]{Fig. 5(a)} illustrates the dynamics consistency error for both FLEXplore and DLPA across six benchmarks. Aside from the \texttt{Goal}, FLEXplore consistently learns dynamics with larger or comparable dynamics consistency errors than those of DLPA. Theoretically, this suggests that FLEXplore has learned looser dynamics. Coupled with the results presented in \hyperref[Section 7.A]{Section 7.A}, we can conclude that FLEXplore has indeed acquired dynamics that are more flexible than those learned by DLPA. The enhanced flexibility enables the agent to explore previously unknown states to a greater extent.

\subsubsection{Adversarial Sample Test}
To intuitively illustrate FLEXplore's capability to learn a flexible dynamics, we develop the Adversarial Sample Test. Specifically, we employ FGSM \cite{goodfellow2014explaining} to generate adversarial samples from the replay buffer for states and continuous parameters, which are reintroduced into the replay buffer at a specified ratio (set to 0.1 in this experiment).  Since the next state remains unchanged, this approach can effectively approximates that introducing adversarial noise to the environment dynamics. Notably, adversarial samples of rewards corresponding to state-action pairs are excluded, making asymptotic performance a reasonable evaluation metric. One intuitive reason the adversarial sample test reflects learned dynamics flexibility is that adversarial noise is added to the environment dynamics only during the training phase. If the learned dynamics lack flexibility—meaning they are overly aligned with the environment dynamics—performance during testing will degrade significantly. Conversely, if the agent learns a flexible dynamics, these issues will not arise. A flexible learned dynamics can capture the overall structure of the environment dynamics without being overly sensitive to details, thereby resisting adversarial noise within a tolerable range.

\hyperref[Fig. 5]{Fig. 5(b)} illustrates the asymptotic performances of FLEXplore and DLPA under adversarial sample test with different strengths, which shows FLEXplore can learn a more flexible dynamics. Strength is defined as the ratio of the L2 norm of the difference between the generated adversarial sample and the original sample to the L2 norm of the original sample. Within the tolerable strength range, FLEXplore exhibits less performance degradation compared to DLPA, as demonstrated on relative simple benchmarks such as \texttt{Platform}, \texttt{Goal}, \texttt{Catch point} and \texttt{Hard move(n=4)}. On complex benchmarks such as \texttt{Hard move(n=6)} and \texttt{Hard move(n=8)}, FLEXplore's asymptotic performance is lower than that of DLPA when the strength exceeds 0.3. We hypothesize that this occurs because the benchmark is overly complex, causing the agent to consistently receive negative reward signals. The addition of adversarial noise, however, encourages the agent to explore to some degree. Such exploration appears necessary because the learned dynamics of DLPA are more closely aligned with the environment dynamics. During training, the agent may venture into unpredictable regions due to the inclusion of adversarial samples.
\begin{figure}\label{Fig. 6}
    \centering
    \includegraphics[width=1\linewidth]{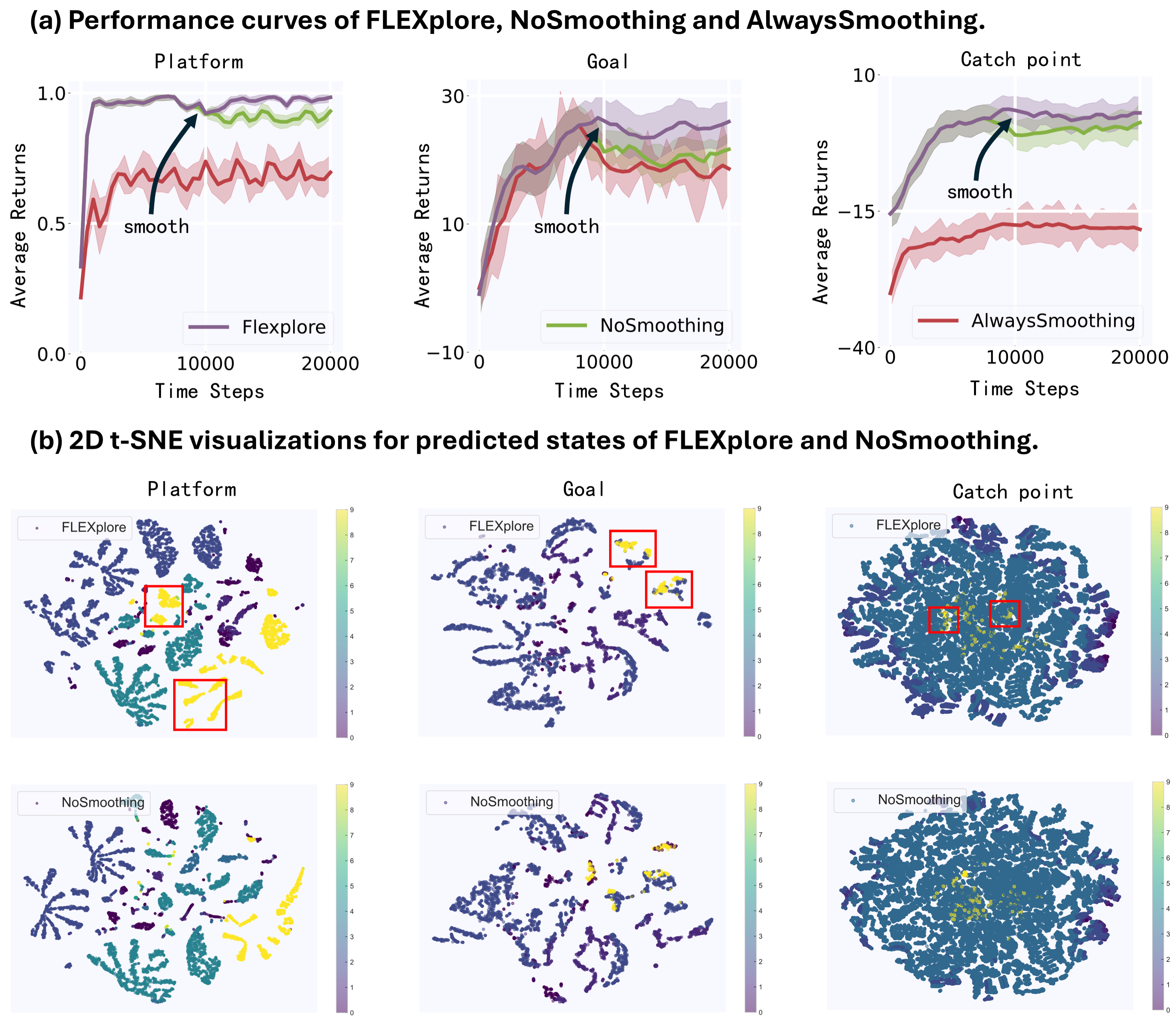}
    \caption{Investigation on reward smoothing mechanism. (a) Performance evaluation of FLEXplore and its ablations, that is, \textit{NoSmoothing} and \textit{AlwaysSmoothing}, over 4 seeds across \texttt{Platform}, \texttt{Goal} and \texttt{Catch point}. (b) 2D t-SNE visualization of environment states. To emphasize the effect of reward smoothing, we specifically delineate FLEXplore's expanded high-reward region with a \textcolor{red}{red} border relative to NoSmoothing.}
    \label{fig:enter-label}
\end{figure}
\subsection{Reward Smoothing} \label{Section 7.C}
In this section, we demonstrate the effectiveness of reward smoothing by evaluating both overall performance and the distribution of rewards.
\subsubsection{Performance}
In \hyperref[Section 4]{Section 4}, we highlight the importance of performing reward smoothing when the training process has reached a stable state, i.e., when the agent demonstrates a reasonable trajectory return. If reward smoothing is applied prematurely during the early phase of training, the agent may theoretically learn an inaccurate reward function. To validate this hypothesis, we include two additional scenarios: \textit{AlwaysSmoothing}, where reward smoothing is applied from the start of training, and \textit{NoSmoothing}, where no smoothing is applied throughout the training. To ensure consistency and eliminate the influence of the training time step threshold $T$ across different environments, we uniformly set $T$ to 10,000 steps.

 \begin{table}[t] \label{Table 2}
    \centering
    \setlength{\tabcolsep}{15pt}
    \caption{The asymptotic performance of FLEXplore's ablations on all the six benchmarks at the end of training.}
    \begin{tabular}{l|c|c}
        \toprule
        ~ & \textbf{NoSmoothing} & \textbf{AlwaysSmoothing} \\
        \midrule
        \textbf{Platform} &  $0.94 \pm 0.04$ & $0.76 \pm 0.09$ \\
        \textbf{Goal} & $30.07 \pm 4.11$ & $28.23 \pm 4.83$ \\
        \textbf{Catch point} & $8.12 \pm 2.32$ & $-10.29 \pm 5.17$ \\
        \textbf{Hard move(4)} & $5.61 \pm 2.31$ & $-12.87 \pm 6.09$ \\
        \textbf{Hard move(6)} & $6.87 \pm 4.89$ & $-34.91 \pm 14.89$ \\
        \textbf{Hard move(8)} & $\mathbf{6.43 \pm 8.22}$ & $-36.28 \pm 13.06$\\ 
        \bottomrule
    \end{tabular}
    \label{tab:my_label}
\end{table}

\hyperref[Fig. 6]{Fig. 6(a)} presents the agent's performance in three relatively simple 
 environments: \texttt{Platform}, \texttt{Goal}, and \texttt{Catch Point}. When reward smoothing is introduced during the stable stage of training, it leads to improved model performance. However, applying reward smoothing early in the training process results in significantly worse agent performance, often underperforming compared to other baselines. Notably, in \texttt{Goal}, the \textit{AlwaysSmoothing} exhibits performance similar to the \textit{NoSmoothing}. This could be attributed to the fact that \texttt{Goal} does not require the agent to learn a highly precise reward function. We observe that the error between the learned reward function and the true reward function in \texttt{Goal} is several orders of magnitude larger than the error observed in the other environments.

We further evaluate the asymptotic performance of FLEXplore and its ablations on six standard PAMDP benchmarks, as shown in \hyperref[Table 2]{Table 2}. In most environments, appropriately applying reward smoothing leads to positive outcomes. However, in the most complex environment, Hard move(8), the effect of reward smoothing is detrimental. This may be due to the high number of negative reward signals in this environment, where even the maximum reward extracted during the stable phase of training is negative. Smoothing such negative reward signals is harmful, as it incentivizes the agent to move towards states associated with negative rewards, making the learning process more challenging. Additionally, we notice that when reward smoothing is applied from the start of training, performance deteriorates significantly. At this phase, the error between the learned reward function and the true reward function is substantial, allowing the agent to mistakenly receive positive rewards for incorrect actions.

\begin{figure*}[t] \label{Fig. 7}
     \centering
     \includegraphics[width=0.9\linewidth]{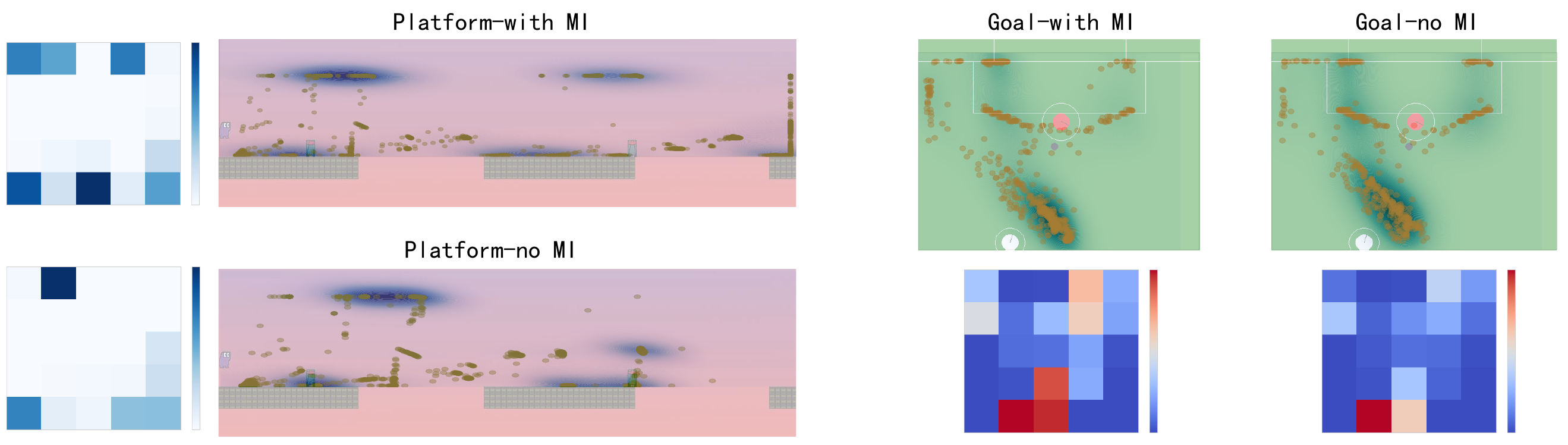}
     \caption{The visualization of the agent/ball's position at each time step on the \texttt{Platform} and \texttt{Goal}. To highlight the enhancement in exploration at the early training phase, we only visualize the agent/ball's movement for the first 2,000 time steps. We use scatter points to mark the agent/ball's position every 5 time steps, allowing the degree of overlap to be inferred from the intensity of the scatter point density. Additionally, heat maps are used to more clearly illustrate the distribution of the agent/ball's positions.}
     \label{fig:enter-label}
 \end{figure*}
 
\subsubsection{Reward Distribution}
A common concern is that the improvement in performance due to reward smoothing, as shown in \hyperref[Fig. 6]{Fig. 6(a)}, does not align with our theoretical analysis. Therefore, we further visualize the distribution of rewards with respect to states.

Note that the state here refers not only to the state of the actionable object, i.e., the player, but also to the state of other objects within the environment. For instance, this could include an obstacle on the platform or a goalie at the goal. When the same state of the player interacts with different states of other objects, the reward may change significantly. Given that the state is represented as a high-dimensional vector, we employ 2D t-SNE visualization \cite{van2008visualizing} to illustrate the distribution of rewards across the state. Specifically, we employ t-SNE for manifold learning by projecting a high-dimensional state vector onto a two-dimensional plane.  Simultaneously, we divide the rewards corresponding to each state into 10 levels, and the state projections on the plane were color-coded according to the reward level assigned.  In this way, we can analyze the effects of reward smoothing by observing the distribution of high-reward areas on the projection map.

The results are shown in \hyperref[Fig. 6]{Fig. 6(b)}. To emphasize that reward smoothing results in a broader range of high-reward regions, we highlight the new high-reward regions of FLEXplore by framing them with a red border, in comparison to \textit{NoSmoothing}. It can be observed that reward smoothing causes the yellow scatter to become increasingly dispersed, indicating the expansion and spread of the high-reward regions. This is consistent with our theoretical design, which emphasizes reward smoothing exclusively for highly rewarding states.

\subsection{Exploration Enhancement} \label{Section 7.D}
In this section, we demonstrate that introducing auxiliary reward signals based on maximizing mutual information enhances the agent's exploration efficiency at the early training phase. As shown in \hyperref[Fig. 7]{Fig. 7}, we train FLEXplore with the auxiliary loss signal (with-MI) and a version without this signal (no-MI) in \texttt{Platform} and \texttt{Goal}, respectively. In \texttt{Platform}, incorporating auxiliary reward signals disperses the agent’s state distribution and increases the success rate. This is evident as the scatter points are more widely distributed, with a greater number of points on the right platform, indicating successful gameplay (the left part). In \texttt{Goal}, it is important to note that the scatter points represent the position of the ball rather than the agent, as the ball's position is crucial to game success and indirectly reflects the agent's movements and actions. Compared to the version without the auxiliary reward signal (no-MI), FLEXplore (with-MI) results in a more dispersed distribution of ball positions. While this dispersion is not immediately apparent in the original scatter plot, it is clearly reflected in the heat map. Additionally, a greater concentration of scatter points along the goal line indicates higher game success rate (the right part).

We can explain this phenomenon in terms of a chain reaction: Since the agent initially adopts near-random policies, directly achieving game success is almost impossible. Without an auxiliary reward signal, the agent may be stuck at specific failure points (e.g., repeatedly getting captured by enemies, falling into a gap in \texttt{Platform}, or being intercepted by the goalkeeper in \texttt{Goal}). While repeated trial and error may work in simple environments, in complex environments, such learning manner becomes impractical. However, auxiliary reward signals based on maximizing mutual information help mitigate this issue by encouraging the agent to select actions that maximize the information gained from subsequent states, thereby reducing repeated failures.

\section{Conclusion}
In this paper, we propose Flexible Model for Effective Exploration (FLEXplore) under Lipschitz parameterized action Markov decision processes to enhance the agent's learning efficiency. FLEXplore adopts a carefully designed loss function to learn a loose yet flexible dynamics, preventing the learned dynamics from becoming overly precise, which inhibits exploration. We theoretically reveal the relationship between the loss function and the Wasserstein Metric of learned and environmental dynamics. To prevent the agent’s policy from collapsing into deterministic then converging to a local minimum during stable training, we introduce reward smoothing to encourage learning around high-reward states and expand the agent’s learning scope. Addtionally, during the model rollout phase, we employ variational lower-bound approximation to derive auxiliary reward signals based on maximizing mutual information to enhance the agents' exploration efficiency. Our experiments demonstrate that FLEXplore's exceptional learning efficiency and superior asymptotic performance compared to baselines.

\appendix
\numberwithin{equation}{subsection}
\section*{Proofs}
\subsection{The necessary definitions and lemmas.} \label{Appendix A}
Before formally proving our theorems, we need to do some preparatory work. First, we provide a definition of the opposite functions set $\mathcal{F}^*$:
\begin{definition}[Opposite Functions Set] \label{Definition A.1}
    \textit{Given a set of functions $\mathcal{F}$, if for any function $f : \mathbb{R}^n \mapsto \mathbb{R}$ in $\mathcal{F}$, there exists a function $g : \mathbb{R}^n \mapsto \mathbb{R}$ in $\mathcal{F}$ such that:}
    \begin{equation}
        f(\mathbf{x}) = -g(\mathbf{x}), \tag{A1}
    \end{equation}
    \textit{we call $\mathcal{F}$ is opposite functions set, denoted as $\mathcal{F}^*$}.
\end{definition}
Based on the definition of opposite functions set, we present and prove the following lemma:
\begin{lemma}
    \textit{For any function $f : \mathbb{R}^n \mapsto \mathbb{R}$ that belongs to the opposite functions set $\mathcal{F}^*$ , and for two points $\mathbf{x}_1, \mathbf{x}_2 \in \mathbb{R}^n$ in the domain, the following equation holds true:}
    \begin{equation}
        \underset{f \in \mathcal{F}^*}{\text{max}}[f(\mathbf{x}_1) - f(\mathbf{x}_2)] = \underset{f \in \mathcal{F}^*}{\text{max}} | f(\mathbf{x}_1) - f(\mathbf{x}_2) |. \tag{A2}
    \end{equation}
\end{lemma}
\begin{IEEEproof}
    We denote the corresponding maximum function for RHS as $f^*$. Since $f^* \in \mathcal{F}^*$, there must exist an opposite function $g^*$ that satisfies the given conditions at the RHS:
    \begin{equation}
        | f^*(\mathbf{x}_1) - f^*(\mathbf{x}_2) | = |g^*(\mathbf{x}_1) - g^*(\mathbf{x}_2)|. \tag{A3}
    \end{equation}
    By the definition of the opposite function, there is
    \begin{equation}
        f^*(\mathbf{x}_1) - f^*(\mathbf{x}_2) = - (g^*(\mathbf{x}_1) - g^*(\mathbf{x}_2)) \tag{A4}
    \end{equation}
    To select the maximum function for LHS, we only need to select the function that ensures it is greater than 0, that is,
    \begin{equation}
        \text{Maximum function} = \left\{
            \begin{aligned}
                & f^*, \quad & \text{if} \  f^*(\mathbf{x}_1) - f^*(\mathbf{x}_2) \geq 0 \\
                & g^*, \quad & \text{if} \ g^*(\mathbf{x}_1) - g^*(\mathbf{x}_2) \geq 0
            \end{aligned}
        \right. \tag{A5}
    \end{equation}
\end{IEEEproof}

~\\
The following lemma asserts that the function $f$ satisfying the Lipschitz condition in \hyperref[eq7]{Equation 7} forms the opposite functions set $\mathcal{F}^*$:
\begin{lemma}
    \textit{The set $\mathcal{F}$ of function $f$ satisfying the Lipschitz condition $K_{d_\mathbb{R},{d_\mathbb{R}}} \leq 1$ is a opposite functions set $\mathcal{F}^*$.}
\end{lemma}
\begin{IEEEproof}
    if a function $f$ satisfying the Lipschitz condition $K_{d_\mathbb{R},{d_\mathbb{R}}} \leq 1$, there exists:
    \begin{equation}
        \begin{aligned}
            \frac{|f(\mathbf{x}_1) - f(\mathbf{x}_2)|}{||\mathbf{x}_1 - \mathbf{x}_2||} & = \frac{|(-f(\mathbf{x}_1)) - (-f(\mathbf{x}_2))|}{||\mathbf{x}_1 - \mathbf{x}_2||} \\
           & = \frac{|g(\mathbf{x}_1) - g(\mathbf{x}_2)|}{||\mathbf{x}_1 - \mathbf{x}_2||} \leq 1.
        \end{aligned} \tag{A6}
    \end{equation}
    where $g$ is the opposite function of $f$. The above formula shows that $g$ also satisfies the Lipstchiz condition $K_{d_\mathbb{R},{d_\mathbb{R}}} \leq 1$, that is, $g \in \mathcal{F}$. By the \hyperref[Definition A.1]{Definition A.1}, $\mathcal{F}$ is a opposite functions set $\mathcal{F}^*$.
\end{IEEEproof}

\subsection{Proof of Theorem 3.1} \label{Appendix B}
\begin{IEEEproof}
The Wasserstein Metric has the following definition:
\begin{equation}
     W(\mu_1,\mu_2) := \underset{j \in \Lambda}{\text{inf}} \int \int j(s_1,s_2)d(s_1,s_2)ds_2ds_1, \tag{B1}
\end{equation}
 where $\Lambda$ denotes the collection of all joint distributions $j$ on $M \times M$ with marginals $\mu_1$ and $\mu_2$ \cite{dynkin1967general}. Sometimes referred to as "Earth Mover's distance", Wasserstein is the minimum expected distance between pairs of points where the joint distribution $j$ is constrained to match the marginals $\mu_1$ and $\mu_2$ \cite{asadi2018lipschitz}. Wasserstein is linked to Lipschitz continuity using duality:
 \begin{equation}
     W(\mu_1,\mu_2) = \underset{f:K_{d_\mathbb{R},{d_\mathbb{R}}} \leq 1}{\text{sup}} \int(f(s)\mu_1(s) - f(s)\mu_2(s))ds. \tag{B2}
 \end{equation}
 This equivalence, known as Kantorovich-Rubinstein duality \cite{villani2009optimal}, lets us compute Wasserstein Metric by maximizing over a Lipschitz set of functions $f:K_{d_\mathbb{R}},{d_\mathbb{R}}$, a relatively easier problem to solve. If we consider the Wasserstein Metric for dynamics $\mathcal{T}$ and $\hat{\mathcal{T}}$, we can derive
 \begin{equation}
     \begin{aligned}
        W(\mathcal{T},\hat{\mathcal{T}}) &= \underset{f:K_{d_\mathbb{R},{d_\mathbb{R}}} \leq 1}{\text{sup}} \int(f(s)\mathcal{T}(s) - f(s)\hat{\mathcal{T}}(s))ds. \\
        &= \underset{f:K_{d_\mathbb{R},{d_\mathbb{R}}} \leq 1}{\text{sup}} \int f(s)\mathcal{T}(s)ds - \int f(s)\hat{\mathcal{T}}(s)ds. \\
        & = \underset{f:K_{d_\mathbb{R},{d_\mathbb{R}}} \leq 1}{\text{sup}} \mathbb{E}_{s \sim \mathcal{T}(s)} [f(s)] - \mathbb{E}_{s \sim \hat{\mathcal{T}}(s)}[f(s)] \geq 0. \\
    \end{aligned} \tag{B3}
 \end{equation} 
Therefore, we derive that:
\begin{equation}
    \begin{aligned}
         & \sum_{t=t_0}^{t_0+H} \gamma^{t-t_0} W(\mathcal{T}(\cdot|s_t,k_t,z_{k_t}), \hat{\mathcal{T}}(\cdot|s_t,k_t,z_{k_t})) \\ 
         & = \sum_{t=t_0}^{t_0+H} \gamma^{t-t_0} \underset{ f_t:K_{d_\mathbb{R},d_\mathbb{R}} \leq 1}{\text{max}} \mathbb{E}_{s_{t+1} \sim \mathcal{T}} \left[ f_t \left( s_{t+1} \right) \right] \\
         & - \mathbb{E}_{ \hat{s}_{t+1} \sim \hat{\mathcal{T}}} \left[ f_t \left( \hat{s}_{t+1} \right) \right] \\
         & = \sum_{t=t_0}^{t_0+H} \gamma^{t-t_0} \underset{ f_t:K_{d_\mathbb{R},d_\mathbb{R}} \leq 1}{\text{max}} \Big| \mathbb{E}_{s_{t+1} \sim \mathcal{T}} \left[ f_t \left( s_{t+1} \right) \right] \\
         & - \mathbb{E}_{ \hat{s}_{t+1} \sim \hat{\mathcal{T}}} \left[ f_t \left( \hat{s}_{t+1} \right) \right] \Big|. \quad \textbf{Lemma A.1 \& A.2} \\
    \end{aligned} \tag{B4}
\end{equation}
Now, we expand and relax our loss function $\mathcal{L}^\text{ex}(\tau;f)$:
\begin{equation}
    \begin{aligned}
        & \mathcal{L}^\text{ex}(\tau;f) \\
        &= \underset{f:K_{d_\mathbb{R},d_\mathbb{R}} \leq 1}{\text{max}}  \sum_{t = t_0}^{t_0 + H} \gamma^{t - t_0} \Big| \mathbb{E}_{s_{t+1}}f(s_{t+1}) - \mathbb{E}_{\tau}f(\hat{s}_{t+1}) \Big| \\
        & \leq \sum_{t=t_0}^{t_0+H} \gamma^{t-t_0} \underset{ f_t:K_{d_\mathbb{R},d_\mathbb{R}} \leq 1}{\text{max}} \Big| \mathbb{E}_{s_{t+1} \sim \mathcal{T}} \left[ f_t \left( s_{t+1} \right) \right] \\
        & - \mathbb{E}_{ \hat{s}_{t+1} \sim \hat{\mathcal{T}}} \left[ f_t \left( \hat{s}_{t+1} \right) \right] \Big| \quad \\
        & = \sum_{t=t_0}^{t_0+H} \gamma^{t-t_0} W(\mathcal{T}(\cdot|s_t,k_t,z_{k_t}), \hat{\mathcal{T}}(\cdot|s_t,k_t,z_{k_t})).
    \end{aligned} \tag{B5}
\end{equation}
\end{IEEEproof}

\subsection{Proof of Theorem 3.2}\label{Appendix C}
\begin{IEEEproof}
\begin{equation}
    \begin{aligned}
        & \mathcal{L}^\text{ex}(\tau;f) \\
        &= \underset{f:K_{d_\mathbb{R},d_\mathbb{R}} \leq 1}{\text{max}}  \sum_{t = t_0}^{t_0 + H} \gamma^{t - t_0} \Big| \mathbb{E}_{s_{t+1}}f(s_{t+1}) - \mathbb{E}_{\tau}f(\hat{s}_{t+1}) \Big| \\
        & \geq \underbrace{ \mathbb{E}_{s_{t_0+1}}\Big[f^*(s_{t_0+1})\Big] - \mathbb{E}_{\tau} \Big[f^*\big(\hat{s}_{t_0+1})\big)\Big] }_{\text{Wasserstein Metric}} \\
        &+ \underbrace{\sum_{t=t_0+1}^{t_0 + H} \gamma^{t-t_0} \Bigg| \mathbb{E}_{s_{t+1}}\Big[f^*(s_{t+1})\Big] - \mathbb{E}_{\tau} \Big[f^*\big(\hat{s}_{t+1}\big)\Big] \Bigg|}_{\text{Non-negative remainder}} \\
         &\geq W(\mathcal{T}(\cdot|s_{t_0},k_{t_0},z_{k_{t_0}}), \hat{\mathcal{T}}(\cdot|s_{t_0},k_{t_0},z_{k_{t_0}})),
    \end{aligned} \tag{C1}
\end{equation}
where $f^*$ meets the following conditions:
\begin{equation}
    f^* = \underset{f:K_{d_\mathbb{R},d_\mathbb{R}} \leq 1}{\text{arg\,max}} \mathbb{E}_{s_{t_0+1}}\Big[f^*(s_{t_0+1})\Big] - \mathbb{E}_{\tau} \Big[f^*\big(\hat{s}_{t_0+1})\big)\Big]. \tag{C2}
\end{equation}
In the same way, we can prove that 
\begin{equation}
   \begin{aligned}
        & \mathcal{L}^\text{ex}(\tau;f) \geq W(\mathcal{T}          
          (\cdot|s_t,k_t,z_{k_t}),\hat{\mathcal{T}}(\cdot|s_t,k_t,z_{k_t})) \\
        & \forall t \in \{t_0,\dots,t_0 + H  \}.
    \end{aligned} \tag{C3} 
\end{equation}
\end{IEEEproof}

\section*{Derivation}
\subsection{Derivation} \label{Appendix D}
The Lipschitz continuity is expressed as:
\begin{equation}
    f:K_{d_{\mathcal{S}},d_{\mathcal{S}}} := \underset{s_1,s_2 \in \mathcal{S}}{\text{sup}}\frac{|| f(s_1) - f(s_2)||}{||s_1 - s_2||} = C. \tag{D1}
\end{equation}
In Equation D1, it is difficult to obtain the expression of $f$ directly, so a neural network is used for fitting it, denoted as $f_{\mathbf{W}}$. 

Consider a simple FFN $\mathbf{W}x+b$, where $\mathbf{W}$ and $b$ is parameter matrix/vector, and $f$ is a activation function. Inspired by Kantorovich-Rubinstein duality \cite{dynkin1967general}, we set $C = 1$. Then we can get
\begin{equation}
    ||f(\mathbf{W}x_1 + b) - f(\mathbf{W}x_2 + b)|| \leq ||x_1 - x_2||, \tag{D2}
\end{equation}
if $x_1$ is close enough to $x_2$, the left side can be expanded with a first-order estimate, and we get
\begin{equation}
    || \frac{\partial f}{\partial x} \mathbf{W}(x_1 - x_2) || \leq ||x_1 - x_2||. \tag{D3}
\end{equation}
In order for the left side not to exceed the right side, the first thing that must be satisfied is that $\frac{\partial f}{\partial x}$ is bounded, and in fact most of the activation functions commonly used in deep learning meet this requirement. For the sake of further derivation, let's call the activation function ReLU, which satisfies $\partial f / \partial x \leq 1$. Now, we focus on $|| \mathbf{W} (x_1, x_2) || \leq ||x_1 - x_2|| $. In fact, this is an inequation of matrix norm,
\begin{equation}
    || \mathbf{W} (x_1, x_2) || \leq ||\mathbf{W}||_2 ||x_1 - x_2||, \tag{D4}
\end{equation}
where $||\mathbf{W}||_2$ is the spectral norm \cite{berezin2014computing}. In this way, we need to limit the spectral norm of parameter matrix to $1$, that is
\begin{equation*}
    ||\mathbf{W}||_2 = 1. \tag{D5}
\end{equation*}
In practice, we construct regularization terms to implement Equation D5:
\begin{equation*}
    \text{min} \, (||\mathbf{W}||_2 - 1)^2.
\end{equation*}

\section*{Additional Details on FLEXplore}
\subsection{Pseudocode} \label{Appendix E}
In Section 7, we provide an experiment of learning efficiency and asymptotic performance  to further verify the validity of our theorem. Below we provide the pseudocode for it.
\begin{algorithm} \label{alg1}
	\caption{FLEXplore} 
	\label{FLEXplore}
	\begin{algorithmic}
		\REQUIRE Model learning horizon $H$, initial learned dynamics $\mathcal{T}_\phi$, initial learned reward function $\mathcal{R}_\psi$, replay buffer $\mathcal{B}$, discrete action network $\pi_\beta$, continuous parameter network $p_\theta$, training timesteps $timestep$, threshold $T$.
		\FOR{rollout\_time $t \leftarrow 0$ \textbf{to} rollout\_horizon}
        \STATE Sample $N$ hybrid action sequences with horizon $H$ by employing $\pi_\beta$ and $p_\theta$.
        \STATE Input these hybrid action sequences and initial state $s_t$ into learned dynamics $\mathcal{T}_\phi$ to get $N$ trajectories $\{\hat{\zeta}\}_{1:N}$.
        \STATE Use \textcolor{orange}{the improved reward signal} (\hyperref[eq23]{Equation 23}) to compute the cumulative return corresponding to each trajectory.
        \STATE Select the trajectories with top-$n$ cumulative returns.
        \STATE Update $\pi_\beta$ and $p_\theta$.
        \ENDFOR
        \STATE Execute the first hybrid $(\hat{k}_{t_0},\hat{z}_{k_{t_0}})$ action of the optimal trajectory.
        \STATE Sample the trajectories derived $\{ s_t,k_t,z_{k_t},r_t,s_{t+1} \}$ from interacting with the environment and put them in replay buffer $\mathcal{B}$.
        \STATE Initial the total model learning loss $\mathcal{L}^{\text{total}}$ as $0$ and max\_reward as $-100$.
        \FOR{learning\_time $t_0$ \textbf{to} $t_0 + H$}
        \STATE Sample the predicted state $\hat{s}_{t+1}$ and the predicted reward $\hat{r}_t$ from learned dynamics $\mathcal{T}_\phi$ and $\mathcal{R}_\psi$ respectively. 
            \IF{\textcolor{orange}{$timestep > T \wedge \hat{r}_t > \text{max\_reward}$}}
            \STATE \textcolor{orange}{Generate the perturbed state $\tilde{s}$} with \hyperref[eq17]{Equation 17}.
            \ENDIF
            \STATE \textcolor{orange}{$\text{max\_reward} = \text{max}(\text{max\_reward},\hat{r}_t)$}
            \STATE \textcolor{orange}{Compute the total dynamics learing loss $\mathcal{L}^\text{total}_{\text{dyn}}$} with \hyperref[eq15]{Equation 15} \textcolor{orange}{and total reward learning loss $\mathcal{L}^\text{total}_{\text{rew}}$} with \hyperref[eq18]{Equation 18}.   
        \ENDFOR
        \STATE $\mathcal{L}^{\text{total}} = \mathcal{L}^\text{total}_{\text{dyn}}+ \alpha \mathcal{L}^\text{total}_{\text{rew}}$
        \STATE Update $\mathcal{T}_\phi$ and $\mathcal{R}_\psi$.
	\end{algorithmic} 
\end{algorithm}
In \hyperref[alg1]{Algorithm 1}, we highlight the main contribution of FLEXplore in \textcolor{orange}{orange font}, which differentiates it from traditional MBRL algorithms.

\subsection{PAMDP Benchmark Environments} \label{Appendix F}
We conduct our experiments on several hybrid action environments, and the detailed experimental description is provided below.
\begin{itemize}
    \item \textbf{Platform} \cite{Masson2016ParamActions}: The agent needs to reach the final goal while evading enemies and avoiding gaps. To accomplish this, the agent selects a discrete action (run, hop, leap) and at the same time determines the corresponding continuous action (horizontal displacement).
    \item \textbf{Goal} \cite{Masson2016ParamActions}: The agent shoots the ball into the gate to win. It has three types of hybrid actions available: kick-to$(x, y)$, shoot-goal-left$(h)$, and shoot-goal-right$(h)$. The continuous action parameters, position$(x, y)$ and position$(h)$ along the goal line, differ significantly.
    \item \textbf{Catch point} \cite{fan2019hybrid}: The agent should catch the target point in limited opportunity (10 chances). There are two hybrid actions \textit{move} and \textit{catch}. \textit{Move} is parameterized by a continuous action value which is a directional variable and \textit{catch} is to try to catch the target point.
    \item \textbf{Hard move} \cite{li2021hyar}: The agent is required to control $n$ equally spaced actuators to reach the target area. It can choose to turn each actuator on or off, leading to an exponential growth in the size of the action set, specifically $2^n$. Each actuator is responsible for controlling movement in its designated direction, and as $n$ increases, the dimensionality of the action space also increases.
\end{itemize}

\subsection{Experimental Details}
Since we use the same network architecture as DLPA, we report only the newly introduced hyperparameters:
\begin{table}[h]
    \centering
    \caption{Hyperparameters}
    \begin{tabular}{l|c|c|c|c|c|c}
        \toprule
        ~ & $\lambda$ & $\mu$ & $\eta$ & $\epsilon$ & $T$ & $H$\\
        \midrule
        \textbf{Platform} & $0.7$ & $0.5$ & $0.01$ & $0.1$ & $10,000$ & $8$\\
        \textbf{Goal} & $0.3$ & $0.5$ & $0.01$ & $0.1$ & $10,000$ & $8$\\
        \textbf{Catch point} & $0.4$ & $0.5$ & $0.01$ & $0.1$ & $10,000$ & $5$\\
        \textbf{Hard move(4)} & $0.4$ & $0.5$ & $0.05$ & $0.3$ & $50,000$ & $5$\\
        \textbf{Hard move(6)} & $0.4$ & $0.5$ & $0.05$ & $0.3$ & $100,000$ & $5$\\
        \textbf{Hard move(8)} & $0.4$ & $0.5$ & $0.05$ & $0.5$ & $100,000$ & $5$\\
        \bottomrule
    \end{tabular}
    \label{tab:my_label}
\end{table}
In practice, we use a three-layer MLP to approximate $f$. The input-dim, hidden-dim and output-dim are state-dim, $32$, $1$ respectively. The activation function is LeakyReLU.

\vspace{11pt}

\vfill

\end{document}